\RequirePackage{fix-cm}
\documentclass[smallextended]{svjour3}       
\smartqed  
\usepackage{graphicx,subfigure}
\usepackage{color}
\usepackage[utf8x]{inputenc}
\usepackage{amsmath}
\usepackage{amssymb}
\usepackage{gensymb}
\usepackage{url}
\usepackage{booktabs}
\usepackage{multirow}
\usepackage{todonotes}
\usepackage{subfigure}

\newenvironment{allintypewriter}{\ttfamily}{\par}

\newcommand{\edition}[1]{{\color{black}{#1}}}

\DeclareMathOperator{\embedding}{embedding}

\begin{document}


\title{Explaining First Impressions: Modeling, Recognizing, and Explaining Apparent Personality from Videos\thanks{$^*$ Means equal contribution by the authors.
}
}

\titlerunning{Explaining first impressions}        

\author{Hugo Jair Escalante$^*$ \and Heysem Kaya$^*$ \and Albert Ali Salah$^*$ \and  Sergio Escalera \and  Ya\u{g}mur G\"u\c{c}l\"ut\"urk \and  Umut G\"u\c{c}l\"u \and Xavier Bar\'{o} \and Isabelle Guyon \and Julio Jacques Junior \and Meysam Madadi \and Stephane Ayache  \and  Evelyne Viegas  \and Furkan G{\"u}rp{\i}nar  \and Achmadnoer Sukma Wicaksana \and Cynthia C.~S. Liem \and Marcel A. J. van Gerven \and Rob van Lier
}

\authorrunning{H.J. Escalante et al.} 

\institute{Hugo Jair Escalante \at
INAOE, Mexico and ChaLearn, USA 
\email{hugojair@inaoep.mx}           
              \and
              Heysem Kaya \at
              Nam{\i}k Kemal University, Department of Computer Engineering, Turkey\\
              \email{hkaya@nku.edu.tr}
              \and
              Albert Ali Salah \at 
              Bo{\u g}azi{\c c}i University, Dept. of Computer Engineering, Turkey and Nagoya University, FCVRC, Japan\\         
              \email{salah@boun.edu.tr} 
              \and
              Furkan G{\"u}rp{\i}nar\at
              Bo{\u g}azi{\c c}i University, Computational Science and Engineering, Turkey\\         \email{furkan.gurpinar@boun.edu.tr} 
\and
Sergio Escalera \at 
University of Barcelona and Computer Vision Center, Spain\\
\email{sergio@maia.ub.es}
\and
Meysam Madadi \at 
Computer Vision Center, Spain\\
\email{mmadadi@cvc.uab.es}
\and 
Yağmur Güçlütürk, Umut Güçlü, Marcel A. J. van Gerven and Rob van Lier \at Radboud University, Donders Institute for Brain, Cognition and Behaviour, Nijmegen, the Netherlands \email{\{y.gucluturk,u.guclu,m.vangerven,r.vanlier\}@donders.ru.nl}
              \and
Xavier Bar\'{o} and Julio Jacques Junior \at Universitat Oberta de Catalunya and Computer Vision Center, Spain\\
\email{\{xbaro,jsilveira\}@uoc.edu}
              \and
Isabelle Guyon \at UPSud/INRIA, Universit\'e Paris-Saclay, France and ChaLearn,  USA \\
\email{guyon@chalearn.org}
\and Stephane Ayache \at Aix Marseille Univ, CNRS, LIF, Marseille, France \email{Stephane.Ayache@lif.univ-mrs.fr}
\and Evelyne Viegas \at Microsoft Research, USA \email{evelynev@microsoft.com}
\and
			  Achmadnoer Sukma Wicaksana, Cynthia C.~S. Liem \at
              Multimedia Computing Group, Delft University of Technology, Delft, The Netherlands\\
              \email{sukmawicaksana@gmail.com, c.c.s.liem@tudelft.nl}              
              }

\date{Received: date / Accepted: date}

\maketitle

\begin{abstract}
Explainability and interpretability are two critical aspects of decision support systems. Within computer vision, they are critical in certain tasks related to human behavior analysis such as in health care applications. Despite their importance, it is only recently that researchers are starting to explore these aspects. This paper provides an introduction to explainability and interpretability in the context of computer vision with an emphasis on looking at people tasks. Specifically, we review and study those mechanisms in the context of first impressions analysis. To the best of our knowledge, this is the first effort in this direction. Additionally, we describe a challenge we organized on explainability in first impressions analysis from video. We analyze in detail the newly introduced data set, evaluation protocol, \edition{proposed solutions} and summarize the results of the challenge. Finally, derived from our study, we outline research opportunities 
that we foresee will be decisive in the near future for the development of the explainable computer vision field. 
\end{abstract}

\section{Introduction}
\label{intro} 

Looking at People (LaP) -- the field of research focused on the visual analysis of human behavior -- has been a very active research field within computer vision in the last decade~\cite{clapresources,LaPSi,DBLP:books/daglib/0028268}. Initially, LaP focused on tasks associated with basic human behaviors that were \emph{obviously} visual (e.g., basic gesture recognition~\cite{HAR97,GER00} or face recognition in restricted scenarios~\cite{FERijcv97,eigenfaces}). Research progress in LaP has now led to models that can solve those initial tasks relatively easily~\cite{DBLP:journals/pami/NeverovaWTN16,deepface}. Instead, attention on human behavior analysis has now turned to problems that are not \emph{visually evident} to model / recognize~\cite{DBLP:conf/mm/ValstarSSEJBSCP13,kaya2014continuous,lopez2016eccv}. For instance, consider the task of assessing personality traits from visual information~\cite{lopez2016eccv}. Although there are methods that can estimate \emph{apparent} personality traits with (relatively) acceptable performance,  model recommendations by themselves are useless if the end user is not confident on the model's \emph{reasoning}, as the primary use for such estimation is to understand bias in human assessors. 

Explainability and interpretability are thus critical features of decision support systems in some LaP tasks~\cite{Escalante:2018:book}. The former focuses on mechanisms that can tell what is the rationale behind the decision or recommendation made by the model. The latter focuses on revealing which part(s) of the model structure influences its recommendations. Both aspects are decisive when applications can have serious implications. Most notably, in health care, security and education scenarios.  

There are models that are explainable and interpretable by their nature, e.g., consider Bayesian networks and decision trees. The model representation, the way its parameters are learned or the manner in which inference is performed gives these models a somewhat \emph{white box} appearance. In fact, explainable and interpretable models have been available for a while for some applications within Artificial Intelligence (AI) and machine learning.  However in computer vision, this aspect is only recently receiving proper attention. This is in large part motivated by the developments on deep learning and its clear dominance across many computer vision tasks. Although such deep models have succeeded at reaching impressive recognition rates in diverse tasks, they are \emph{black box models}, as one cannot say too much on the way these methods make recommendations or on the structure of the model itself\footnote{Please note that there have been efforts since a while trying to demystify the structures of deep networks, see e.g.,~\cite{Yosinski15}.}. This perception is starting to change, with more research focused on visualizing and understanding the structure of models and with cutting edge research focused on explaining the recommendations of LaP models. 

This paper comprises a comprehensive study on explainability and interpretability in the context of computer vision, with emphasis on LaP tasks. In particular, we focus on those mechanisms in the context of first impressions analysis. The contributions of this paper are as follows. We review concepts and the state of the art on the subject. We describe a challenge we organized on explainability in first impressions analysis from video. We analyze in detail the newly introduced data set, the evaluation protocol, and summarize the results of the challenge. \edition{Top ranked solutions to the challenge are described  and evaluated in detail. }
Finally, derived from our study, we outline research opportunities that we foresee will be decisive in the near future for the development of the explainable computer vision field. 


The remainder of this paper is organized as follows. Section 2 reviews related work on explainability and interpretability in the context of computer vision. 
Section~\ref{sec:challenge} describes the LAP First Impression Challenge, summarizing its results and main findings. Section~\ref{sec:solutions_from_participants} describes in more detail methodologies proposed for explaining first impressions. Section~\ref{sec:analisys_data} presents an in-depth study on the data set associated with the challenge. Finally, Section~\ref{sec:lessons_learned} presents a discussion on the lessons learned and outlines further research ideas in this area. 











\section{Explainability and interpretability in computer vision}


This section reviews related work on first impressions analysis and explainable models in LaP and computer vision. 

\subsection{Audio-visual analysis of First impressions}

LaP has pushed the state of the art in classical problems that have strong visual aspects, such as face recognition, body pose estimation, and gesture recognition.
However, there are several problems for which research is still in its infancy. 
In this paper we approach the problem of estimating the apparent personality of people and related variables, see e.g.~\cite{lopez2016eccv}. Personality and conduct variables in general are rather difficult to infer precisely from visual inspection, this holds even for humans. Accordingly, the LaP field is starting to pay attention to a less complex problem, that of estimating \emph{apparent} personality from visual data~\cite{lopez2016eccv}. Related topics receiving increasing attention from the LaP community are first impressions analysis, depression recognition and hiring recommendation systems~\cite{lopez2016eccv,DBLP:conf/mm/ValstarSSEJBSCP13,icpr_contest,GatiaPerez:2013}, all of them starting from visual information. \edition{For a comprehensive review on apparent personality analysis, from a computer vision point of view, we refer the reader to~\cite{Jacques:Arxiv:2018}.}

The impact that LaP methods can have in such problems is huge, as \edition{... \emph{any technology involving understanding, prediction and synthesis of human behavior is likely to benefit from personality computing approaches...}~\cite{Vinciarelli:TAC2013}}. One of such application is \emph{job candidate screening}. According to Nguyen and Gatica-Perez~\cite{Nguyen2016}, video interviews are starting to modify the way in which applicants get hired. The advent of inexpensive sensors and the success of online video platforms has enabled the introduction of a sort of video-based resum\'e. In comparison with traditional document-based resum\'es, video-based ones offer the possibility for applicants to show their personality and communication skills. If these sort of resum\'es are accompanied by additional information (e.g., paper resum\'e, essays, etc.), recruitment processes can benefit from automated job screening in some initial stages. But more importantly, assessor bias can be estimated with these approaches, leading to fairer selection. On the side of the applicant, this line of research can lead to effective coaching systems to help applicants present themselves better and to increase their chances of being hired. \edition{This is precisely the aim of the speed interviews project\footnote{\url{http://gesture.chalearn.org/speed-interviews}} that gave origin to the present study.}  

Efforts on automating the interview process by analyzing videos are scarce~\footnote{However, one should note that the analysis of non-verbal behavior to predict the outcome of a social interaction is a topic that has been studied for a while in different domains~\cite{GatiaPerez:2013}.}.  
In~\cite{Nguyen2016}
the formation of job-related first impressions in online conversational audiovisual resum\'es is analyzed. Feature representations are extracted from audio and visual modalities. Then, linear relationships between nonverbal behavior and the organizational constructs of ``hirability'' and personality are examined 
via correlation analysis. 
Finnerty et al.~\cite{Finnerty:2016} aimed to determine whether first impressions of stress (from an annotated database of job interviews) are equivalent to physiological measurements of electrodermal activity (EDA). In their work, automatically extracted nonverbal cues, stemming from both the visual and audio modalities were examined.  Then, two regression techniques, ridge regression and random forest are evaluated.
Stress impressions were found to be significantly negatively correlated with ``hirability'' ratings (i.e., individuals who were perceived to be more stressed were more likely to obtain lower ``hirability'' scores). Regression results show evidence that visual features are better predictors of stress impressions than audio features.
In the same line, Naim et al.~\cite{Naim:2016} exploited verbal and nonverbal behaviors in the context of job interviews from face to face interactions. Their approach includes facial expression, language and prosodic information analysis. The framework is capable of making recommendations for a person being interviewed, so that he/she can improve his/her ``hirability ''score based on the output of a support vector regression model.

\subsection{Explainability and interpretability in the modeling of visual information}

Following the great success obtained by deep learning based architectures in recent years, different models of this kind have been proposed to approach the problem of first impression analysis from video interviews/resum\'es or video blogs (including related tasks such as job recommendation)~\cite{Gucluturk2017,Gucluturk_2017_ICCV,Ventura:CVPRW2017}. Although very competitive results have been reported with such methods (see e.g.,~\cite{DBLP:conf/ijcnn/EscalanteGEJMBA17}), a problem with such models is that 
they are often perceived as \textit{black-box techniques}: they are able to effectively model very complex problems, but they cannot be interpreted, nor can their predictions be explained~\cite{Ventura:CVPRW2017}. 
Because of this, explainability and interpretability have received special attention in different fields, see e.g.,~\cite{darpa}. In fact, the interest from the community on this topic is evidenced by the organization of dedicated events, such as thematic workshops~\cite{WHI2016,WHI2017,NIPS:WKSyM2017,WIML:NIPS16,NIPS:INSyM2017} and challenges~\cite{DBLP:conf/ijcnn/EscalanteGEJMBA17}. This is particularly important to ensure fairness and to verify that the models are not plagued with various kinds of biases, which may have been inadvertently introduced.

%


Among the efforts for making models more explainable/interpretable, visualization  
has been seen as a powerful technique to understand how deep neural networks work~\cite{Kindermans:2016,Zeiler2014,Mahendran:2016,ZintgrafCAW17,Wang2017}. These approaches primarily seek to understand what internal representations are formed in the black box model.
Although visualization by itself is a convenient formulation to understand model structure, approaches going one step further can also be found in the  literature\edition{~\cite{Das:CVIU2017,gradcam2016,gradcam2016:full,Wei-Koh:ICML2017,Hendricks:ECCV:2016,Kim:ICCV:2017}}. 
Selvaraju et al. presented a technique for making convolutional neural network (CNN) based models more transparent~\cite{gradcam2016,gradcam2016:full}. 
A novel class-discriminative localization technique is proposed 
and combined with existing
high-resolution visualizations to produce visual explanations for CNN-based models. 

Das et al. conducted large-scale studies on ``human attention'' in Visual Question Answering (VQA) in order to understand where humans look when answering questions about images~\cite{Das:CVIU2017}. Attention maps generated by deep VQA models were evaluated against human attention. 
Their experiments showed that current attention models in VQA do not seem to be looking at the same regions as humans, \edition{although maps brought  light on how the model analyzes images}. 
\edition{In the same context, Rajani and Mooney combined multiple VQA models to obtain better explanations (maps)~\cite{rajani2018}, and the assembled explanations proved to improve those of individual models.}


\edition{In~\cite{Hendricks:ECCV:2016}, visual explanations for image classification are generated by focusing on the discriminating properties of visible objects. The proposed approach jointly predicts a class label and explains why the predicted label is appropriate for the image. The explanation mechanism resembles an image captioning method. 
More recent work in the same line can be found in~\cite{akata2018}. 
In the context of self-driving cars, Kim and Canny~\cite{Kim:ICCV:2017} presented a visual explanation method taking the form of real-time highlighted regions of an image that ``causally'' influence the network's output. A visual attention model is used in a first stage to highlight image regions that potentially influence the output. Then, a causal filtering step is applied to determine which input regions actually influence the output. We refer the reader to~\cite{Escalante:2018:book} for a compilation on recent progress  explainability and interpretability in the context of Machine Learning and Computer Vision. 
}

\edition{Although this paper is focused on visual content-based analysis, it is important to note that expainability and interpretability for audio~\cite{Becker:2018:arxiv} and/or text information~\cite{Arras:2017:PlosOne} also have attracted the attention from the Machine Learning and Natural Language Processing communities. A more general review on explainability and justification in the Machine Learning and adjacent fields has been presented in~\cite{Biran:2017:IJCAIW}. However, in their work, expainability and interpretability with respect to deep learning-based models have not been discussed in depth. }


\subsection{Explainability and interpretability of first impressions}
Methods for first impressions analysis developed so far are limited in their explainability and interpretability capabilities.
The question of why a particular individual receives a positive (or negative) evaluation deserves special attention, as such methods will influence our lives strongly, once they become more and more common. Recent studies, including those submitted to a workshop we organized - ChaLearn: Explainable Computer Vision Workshop and Job Candidate Screening Competition at CVPR2017\footnote{\url{http://openaccess.thecvf.com/CVPR2017_workshops/CVPR2017_W26.py}}, sought to address this question. In the remainder of this section, we review these first efforts on explainability and interpretability for first impressions and ``hirability'' analyses.  

G\"u\c{c}l\"ut\"urk et al.~\cite{Gucluturk2017} proposed 
a deep residual network, trained on a large dataset of short YouTube video blogs, for predicting first impressions and whether persons seemed \emph{suitable} to be invited to a job interview. In their work, they use a linear regression model that predicts the interview annotation (``invite for an interview'') as a function of personality trait annotations in the five dimensions of the Big-Five personality model. The average ``bootstrapped'' coefficients of the regression are used to assess the influence of the various traits on hiring decisions. The trait annotations were highly predictive of the interview annotations ($R ^ 2 = $ 0.9058), and the predictions were significantly above chance level ($p \ll 0.001$, permutation test). Conscientiousness had the largest and extroversion had the smallest contributions to the predictions ($\beta > 0.33$ versus $\beta < 0.09$, respectively). For individual decisions, the traits corresponding to the two largest contributions to the decision are considered ``explanations''. In addition, a visualization scheme based on representative face images was introduced to visualize the similarities and differences between the facial features of the people that were attributed the highest and lowest levels of each trait and interview annotation. 

In~\cite{Gucluturk_2017_ICCV}, the authors identified and highlighted  the audiovisual information used by their deep residual network through a series of experiments in order to explain its predictions. Predictions were \textit{explained} using different strategies, based either on the visualization of representative face images~\cite{Gucluturk2017}, or using an audio/visual occlusion based analysis. The later involves systematically masking the visual or audio inputs to the network while measuring the changes in predictions as a function of location, predefined region or frequency band. This approach marks the features to which the decision is sensitive (parts of the face, pitch, etc.)


Ventura et al.~\cite{Ventura:CVPRW2017} presented a deep study on understanding why CNN models are performing surprisingly well in automatically inferring first impressions of people talking to a camera. Although their study did not focus on ``hirability'' systems, results show that the face provides most of the discriminative information for personality trait inference, and the internal CNN representations mainly analyze key face regions such as eyes, nose, and mouth. 

Kaya et al.~\cite{Kaya_2017_CVPR_Workshops} described an end-to-end system for explainable automatic job candidate screening from video interviews. In their work, audio, facial and scene features are extracted. Then, these multiple modalities are fed into modality-specific regressors in order to predict apparent personality traits and ``hirability'' scores. The base learners are stacked to an ensemble of decision trees to produce quantitative outputs, and a single decision tree, combined with a rule-based algorithm produces interview decision explanations based on quantitative results. Wicaksana and Liem~\cite{WicaksanaCVPRW:2017} presented a model to predict the Big Five personality trait scores and interviewability of vloggers, explicitly targeting
explainability of the system output to humans without technical background. In their work, multimodal feature representations are constructed to capture facial expression, movement, and linguistic information. These two approaches are discussed in detail in Section~\ref{sec:solutions_from_participants}, as their proposed methods obtained the best performance in the Job Candidate Screening Competition we organized~\cite{Kaya_2017_CVPR_Workshops,WicaksanaCVPRW:2017}. 

\subsection{A word of caution}
The previous sections have reviewed research progress on LaP focusing on the explainability and interpretability of models. Researchers in LaP have made a great progress in different areas of LaP, as a result of which, human-level performance has almost been achieved on a number of tasks (e.g., face recognition) for controlled settings and adequate training conditions. However, most progress has concentrated on obviously visual problems. More recently, LaP is targeting problems that deal with subjective assessments, such as first impression estimation. Such systems can be used for understanding and avoiding bias in human assessment, for implementing more natural behaviors, and for training humans in producing adequate social signals. Any task related to social signals in which computers partake in the decision process will benefit from accurate, but also explainable models. Subsequently, this line of research should not be conceived of implementing systems that may (in some dystopic future) dislike a person's face and deny them a job interview, but rather look at the face and explain why 
the biased human assessor denied the job interview.

\section{The job candidate screening coopetition}
\label{sec:challenge}
With the goal of advancing research on explainable models in computer vision, we organized an academic \edition{\emph{coopetition}} 
on 
explainable computer vision and pattern recognition to assess ``first impressions" on personality traits. It is called a ``coopetition," rather than a competition, because it promoted code sharing between the participants.
This section describes the challenge and in the next sections we elaborate on the associated data set and main findings. The design of the challenge is further detailed in~\cite{DBLP:conf/ijcnn/EscalanteGEJMBA17}. 

\subsection{Job candidate screening: perspectives from organizational psychology}
\label{sec:firstimpr_review}
The 2017 ChaLearn challenge at CVPR was framed in the context of Job Candidate screening. More concretely, the main task of the challenge was to guess the \edition{\emph{apparent}} first impression judgments on people in video blogs, and whether they will be considered to be invited to a job interview. Accordingly, in this section we briefly review relevant aspects of organizational psychology on job candidate screening. 

Traditionally, job candidate screening and application sifting would be conducted based on information in CVs and application forms, supported by references. The sifting procedure can be further improved by assessing behavioral competences, weighted application blanks and bio-data, training and experience ratings, minimum qualifications, background investigations or positive vetting, structured questioning, Internet tests, and application scanning software~\cite{Cook2009}. 

For personnel selection, various types of assessment may further be performed to assess the candidate's suitability. Seven main aspects are identified in \cite{Cook2009}:
\begin{itemize}
\item{Mental ability (GMA) (intelligence, problem solving, practical judgement, clerical ability, mechanical comprehension, sensory abilities)}
\item{Personality traits}
\item{Physical characteristics (strength, endurance, dexterity)}
\item{Interests, values and fit}
\item{Knowledge (declarative: facts, procedural: knowing how to act, tacit: `knowing how things really happen') (note: mastery of higher-level knowledge may require higher levels of mental ability)}
\item{Work skills (dedicated skills required for the job, e.g. bricklaying or diagnosing an illness)}
\item{Social skills (e.g. communication, persuasion, negotiation, influence, leadership, teamwork).}
\end{itemize}

All of these aspects were considered for the design of the challenge. However, in order to limit its scope and to facilitate the objective evaluation of methods for automatically job candidate screening, some aspects were simplified: \edition{we restrict the challenge to \emph{apparent} personality and to make interview recommendations based solely on appearance. This setting could inherit biases from the data labeling process (see Section~\ref{sec:datadescription}). However, one should note that a job recruiter could have access to the same information available in this challenge.} Obviously, different requirements are required from, say, an HR specialist and a programmer. 
We focused on the aspects that are independent from the job type to obtain general results.
\subsection{Overview}\label{sec:overview}
\begin{figure}[h!tb!] 
    \centering    
    \includegraphics[scale=0.2]{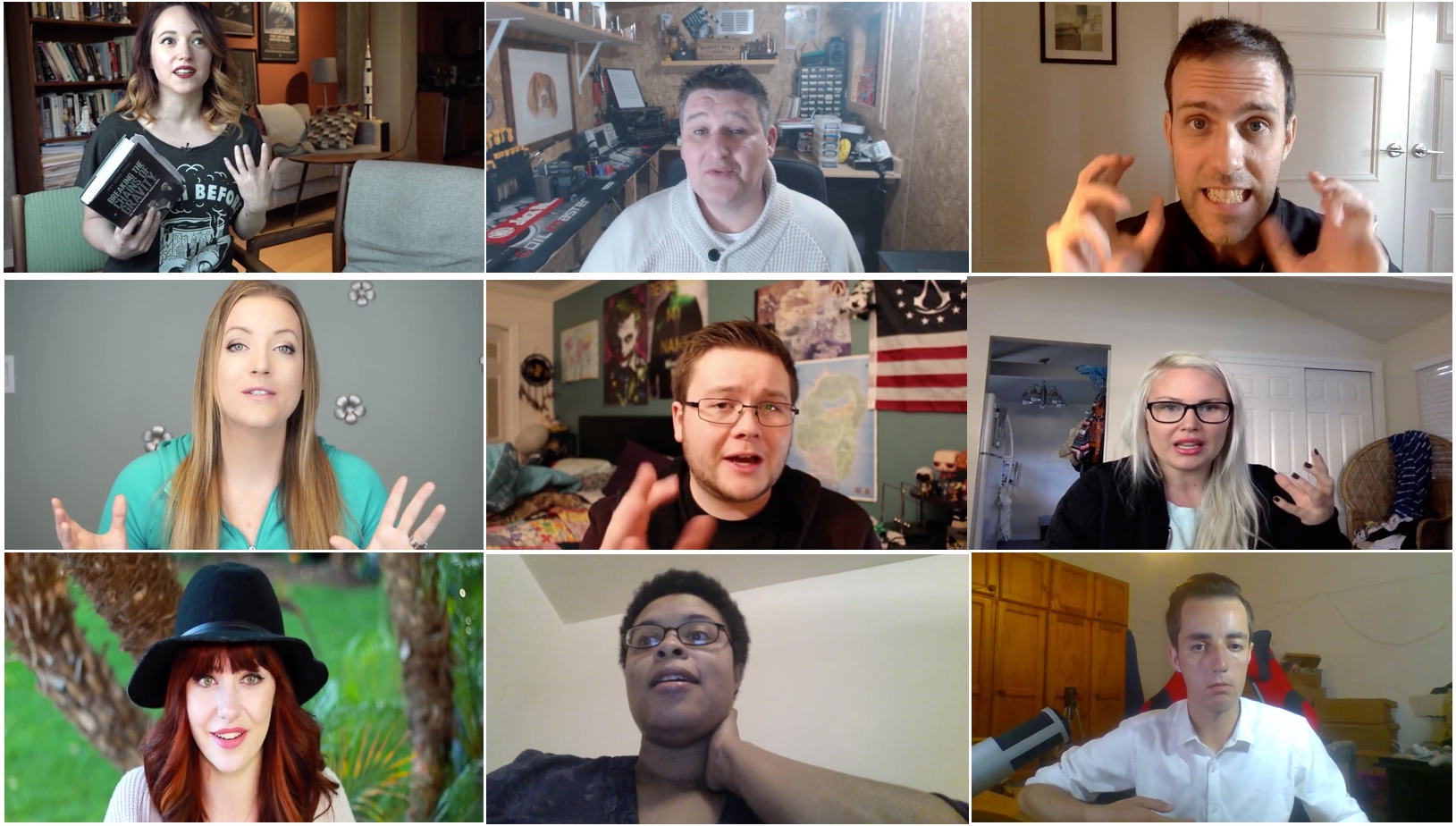}    
    \caption{Snapshots of sample videos from the First Impressions data set~\cite{lopez2016eccv}.}\label{fig:samplesdataset}
\end{figure}

The challenge relied on a novel data set that we  made publicly available recently\footnote{Data set is available at \url{http://chalearnlap.cvc.uab.es/dataset/24/description/}}~\cite{icpr_contest,lopez2016eccv}. The so-called first impressions data set
comprises 10,000 clips (with an average duration of 15s) extracted from more than 3,000 different YouTube high-definition (HD) videos of people facing a camera and speaking in English. People in videos have different gender, age, nationality, and ethnicity (see Section~\ref{sec:analisys_data}). Figure~\ref{fig:samplesdataset} shows snapshots of sample videos from the data set. 

In the coopetition, we challenged the participants to provide predictive models with explanatory mechanisms. The recommendation that models had to make was on whether a job candidate should be invited for an interview or not, by using short video clips (see Sec.~\ref{sec:datadescription}). Since this is a decisive recommendation, we thought explainability would be extremely helpful in a scenario in which human resources personnel wants to know what are the reasons of the model for making a recommendation.  We assumed that the candidates have already successfully  passed technical screening interview steps e.g. based on a CV review. We addressed the part of the interview process related only to {\bf human factors}, complementing aptitudes and competence, which were supposed to have been separately evaluated. Although this setting is simplified, the challenge was a real and representative scenario where explainable computer vision and pattern recognition is highly needed: \emph{a recruiter needs an explanation for the recommendations made by a machine}. 

The challenge was part of a larger project on speed interviews: \url{http://gesture.chalearn.org/speed-interviews}, whose overall goal is to help both recruiters and job candidates by using automatic recommendations based on multi-media CVs. Also, this challenge was related to two previous 2016 competitions on first impressions that were part of the contest programs of ECCV2016~\cite{lopez2016eccv}  and ICPR2016~\cite{icpr_contest}. Both previous challenges focused on predicting the apparent personality of candidates in video. In this version of the challenge, we aimed at predicting {\bf hiring recommendations} in a candidate screening process, i.e. whether a job candidate is worth interviewing (a task not previously explored). More importantly, we focused on the explanatory power of techniques: \emph{solutions have to ``explain" why a given decision was made.}  Another distinctive feature of the challenge is that it incorporates a collaboration-competition scheme (coopetition) by rewarding participants who share their code during the challenge, weighting rewards with the usefulness/popularity of their code. 

\subsection{Data Annotation}
\label{sec:datadescription}
Videos were labeled both with apparent personality traits and a``\emph{job-interview variable}". The considered personality traits were those from the  Five Factor Model (also known as the ``Big Five'' or OCEAN traits)~\cite{McCrae:1992}, which is the dominant paradigm in personality research. It models human personality along five dimensions: \emph{\textbf{O}penness, \textbf{C}onscientiousness, \textbf{E}xtroversion, \textbf{A}greeableness, \textbf{N}euroticism}, respectively. Thus, each clip has ground truth labels for these five traits. Because ``Neuroticism'' is the only negative trait, we replaced it by its opposite (non-Neuroticism) to score all traits in a similar way on an positive scale. Additionally, each video was labeled with a variable indicating whether the subject should be invited to a job interview or not (the ``\emph{job-interview variable}"). 

\begin{figure}[htb] 
    \centering    
    \includegraphics[scale=0.18]{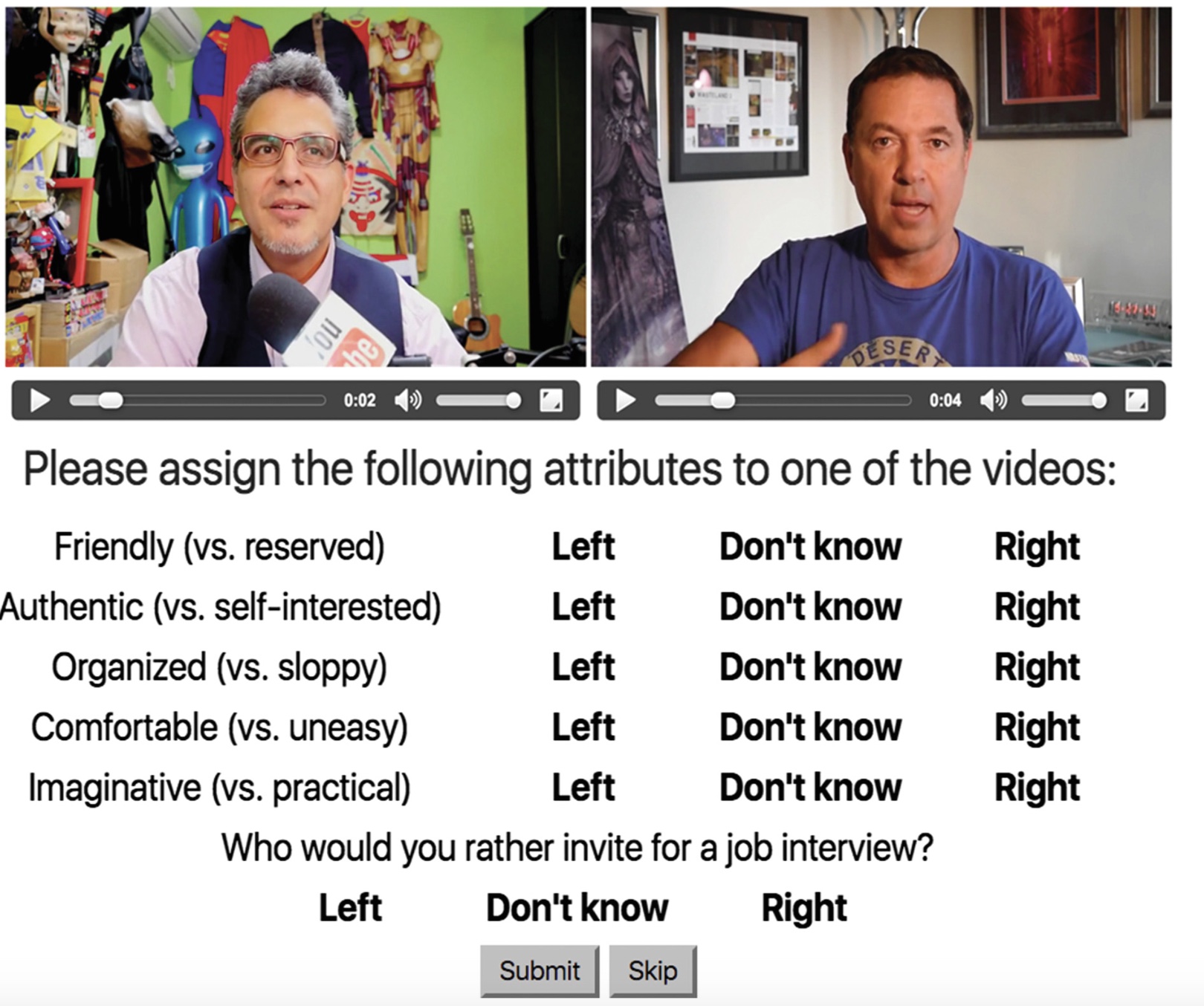} 
    \caption{Snapshots of the interface for labeling videos~\cite{lopez2016eccv}. The ``big five'' traits are characterized by adjectives: Extroversion = Friendly (vs. Reserved); Agreeableness = Authentic (vs. Self-interested); Conscientiousness = Organized (vs. Sloppy); (non-)Neuroticism = Comfortable (vs. Uneasy); Openness = Imaginative (vs. Practical). }\label{fig:amtinterface}
\end{figure}

Amazon Mechanical Turk (AMT) was used for generating the labels. To avoid calibration problems, we adopted a pairwise ranking approach for labeling the videos: each Turker was shown two videos and asked to answer which of the two subjects present individual traits more strongly. Also, annotators were instructed to indicate which of two subjects they would invite for a job interview. In both cases, a neutral, ``I do not know" answer was possible. \edition{During labeling, different pairs of videos were given to different and unique annotators. Around $2500$ annotators labelled the data, and a total of $321,684$ pairs were used~\cite{Chen2016}. Although this procedure does not allow us to perform the agreement analysis among annotators which labeled the same pairs, below we report an experiment that aims at assessing the consistency of labellings from AMT workers and human annotators in a more controlled scenario. 
} Figure~\ref{fig:amtinterface} illustrates the interface that AMT workers had access to. 

We post-processed the rankings provided by the annotators to generate  scores for each annotated variable. By doing so, cardinal scores were obtained by pairwise fitting a Bradley-Terry-Luce (BTL) model. Additionally, we used a small-world algorithm for sampling the pairs that workers had to annotate. Details on the labeling procedure can be found in~\cite{Chen2016}. 

In addition to the audio visual information available in the raw clips, we provided transcripts of the audio.
In total, this added about $375,000$ transcribed words for the entire data set. The transcriptions were  obtained by using a professional human transcription service\footnote{\url{http://www.rev.com}} to ensure maximum quality for the ground truth annotations. 

The feasibility of the challenge annotations was  successfully evaluated prior to the start of the challenge. \edition{The reconstruction accuracy of all annotations obtained by the BTL model was greater than $0.65$ (test accuracy of cardinal rating reconstruction by the model~\cite{Chen2016}).} Furthermore, the apparent trait annotations were highly predictive of invite-for-interview annotations, with a significantly above-chance coefficient of determination of 0.91.

\begin{figure}[htb] 
    \centering    
    \includegraphics[scale=0.5]{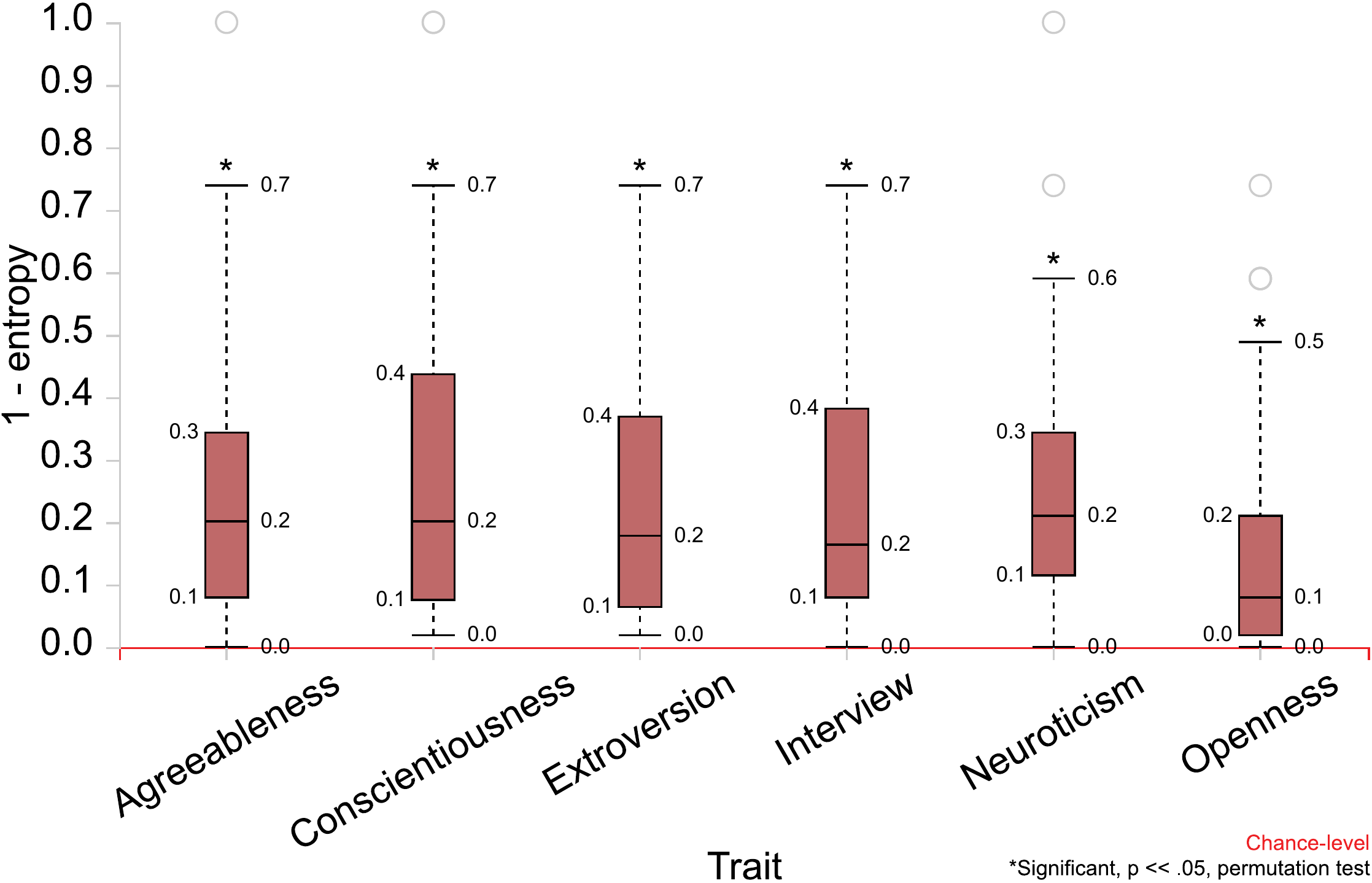} 
    \caption{\edition{Consistency estimates of the dataset per each personality trait. For clarity, 1 - entropy of the distribution of the choices of the participants averaged per each personality trait is displayed, i.e. low values mean low consistency and high values mean high consistency. Note that the consistency of the evaluations for all traits were significantly above the chance level.} }\label{fig:consistency}
\end{figure}

\edition{\subsubsection{Annotation agreement}}
\edition{Since no video pair was viewed by more than 1 person in the original data collection experiment, in order to estimate the consistency of the personality assessments in the dataset we ran a second experiment with 12 participants (6 males and 6 females, mean age = 27.2). This experiment was in most aspects a replication of the original experiment, the main differences being that the same videos were viewed and assessed by all participants and that it was not an online study. The experiment consisted of viewing a subset of 100 video pairs that were randomly drawn from the original dataset, and then making judgements regarding the personality of the people in the videos by answering questions like: “Which person is more friendly?”, etc. similar to the original experiment. They could response by selecting “left”, “right”, or “don’t know” for each pair. Since all participants evaluated the same video pairs in this second experiment, we were able to quantify the consistency of their choices amongst each other.
To measure consistency, for each video pair, we calculated the entropy of the distribution of the choices of the participants, and averaged the results per each personality trait. Entropy can take values between 0 and 1, where a low average entropy value represents high consistency for that trait and a high average entropy value represents low consistency. Note that the mapping between consistency and entropy of a distribution is not linear. Our analysis revealed that all traits were significantly more consistent than chance-level (p $<<$ 0.05, permutation test), with organizedness evaluations that represented conscientiousness trait being the most consistent (entropy = 0.72), and imaginativeness evaluations that represented the openness trait being the least consistent (entropy = 0.85) among participants (See Figure~\ref{fig:consistency}).
}

\edition{First impression annotation is a very complex and subjective task. Several aspects can influence the way people perceive others, such as cultural aspects, gender, age, attractiveness, facial expression, among others (from the observer point of view as well as from the perspective of the person being observed).  It means that, e.g., different individuals can have very distinct impressions of the same person in an image. Moreover, the same individual can perceive the same person differently at different circunstancies (e.g., time intervals, images or videos) due to many reasons. Subjectivity in data labeling, and more specifically in first impression, is a very challenging task which has attracted a lot of attention by the machine learning and computer vision communities. For a comprehensive review on this topic, we refer the reader to~\cite{Jacques:Arxiv:2018}.}

\subsection{Evaluation protocol}

The job candidate screening challenge was divided into two tracks/stages, comprising quantitative and qualitative variants of the challenge. The qualitative track being associated to the explainability capabilities of the models developed for the first track.  
The tracks were run in series as follows: 
\begin{itemize}
\item \textbf{Quantitative competition (first stage).} Predicting whether the candidates are promising enough that the recruiter wants to invite him/her to an interview.

\item \textbf{Qualitative coopetition (second stage).} Justifying/explaining with an appropriate user interface the recommendation made such that a human can understand it. Code sharing was expected at this stage. 
\end{itemize}

\begin{figure*}[htb]
    \centering
  \includegraphics[width=1.0\textwidth]{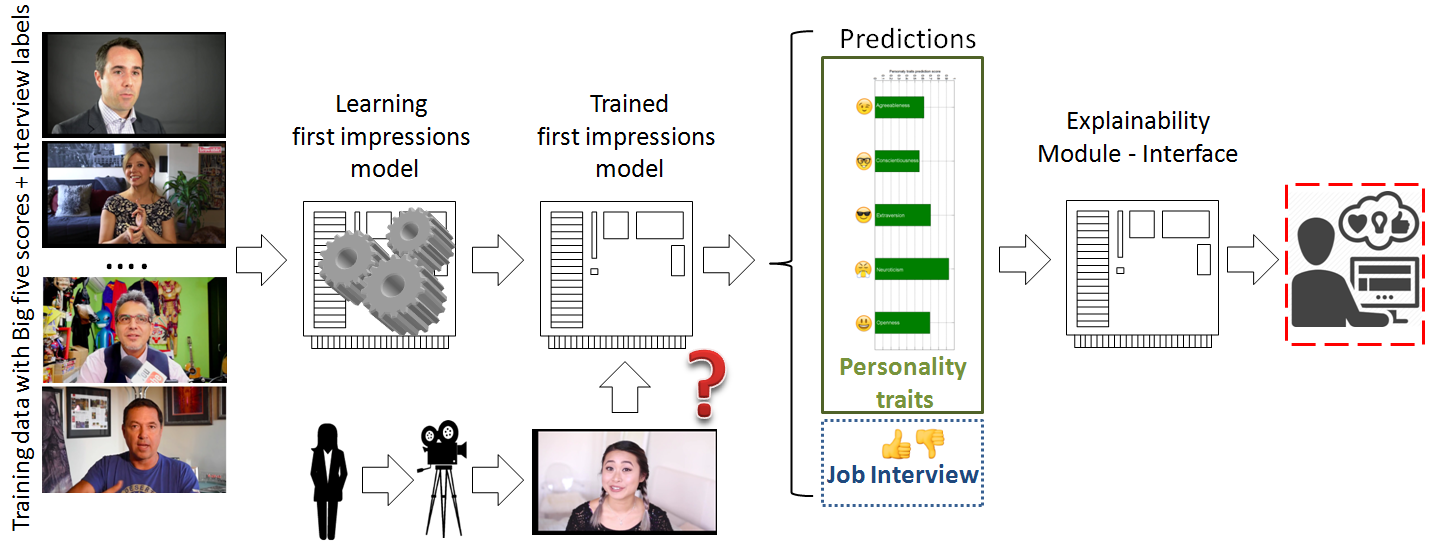}
  \caption{Diagram of the considered scenario in the job candidate screening coopetition. The solid (green) top square indicates the variables evaluated in past editions of the challenge~\cite{icpr_contest,lopez2016eccv}. The dotted (blue) bottom square indicates the variable evaluated in the quantitative track. The dashed (red) square indicates what is evaluated in the qualitative track.}
  \label{fig:escenario}
\end{figure*}

Figure~\ref{fig:escenario} depicts the information that was evaluated in each stage. In both cases, participants were free (and encouraged) to use information from apparent personality analysis. However, please note that the personality traits labels were provided {\em only with training data}. This challenge adopted a \emph{coopetition} scheme; participants were expected to share their code and use other participants's code, 
mainly for the second stage of the challenge: e.g., a team could participate only in the qualitative competition using the solution of another participant in the quantitative competition. 
\subsubsection{Platform}
As in other challenges organized by ChaLearn\footnote{\url{http://chalearn.org}}, the job candidate screening coopetition ran  in  CodaLab\footnote{\url{http://codalab.org/}}; a
platform developed by Microsoft Research and Stanford University in close collaboration with the organizers of the challenge. 
\subsubsection{Data partitioning}
For the evaluation, the data set was split as follows:
\begin{itemize}
\item \textbf{Development (training)} data with ground truth for all of the considered variables (including personality traits) was made available at the beginning of the competition. 

\item \textbf{Validation} data {\bf without labels} (neither for personality traits nor for the ``job-interview variable") was also provided to participants at the beginning of the competition.  Participants could submit their predictions on validation data to the CodaLab platform and received immediate feedback on their performance.  
\item \textbf{Final evaluation (test)} unlabeled data was  made available to participants one week before the end of the quantitative challenge. Participants had to submit their predictions in these data to be considered for the final evaluation (no ground truth was released at this point). 
Only five test set submissions were allowed per team.
\end{itemize}

In addition to submitting predictions for test data, participants desiring to compete for prizes submitted their code for verification, together with fact sheets summarizing their solutions. 

\subsubsection{Evaluation measures}
\label{subsec:evalmeas}
For explainability, qualitative assessment is crucial. Consequently, we provide some detail about our approach in this section. The competition stages were independently evaluated, as follows:

\begin{itemize}
\item \textbf{Quantitative evaluation (interview recommendation).}  The performance of solutions was evaluated according to their  ability  for predicting  the interview variable in the test data. 
Specifically, similar in spirit to a regression task, the evaluation consists in computing the accuracy  over the invite-for-interview variable,  defined as:
\begin{equation} \label{eq:acc}
A = 1 - \frac{1}{N_t} \sum_{i=1}^{Nt} |t_i - p_i|/\sum_{i=1}^{Nt} |t_i -\overline{t}|
\end{equation}
where $p_i$ is the predicted score for sample $i$, $t_i$ is the corresponding ground truth value, with the sum running over $N_t$ test videos, and $\overline{t}$ is the average ground truth score over all videos.

\item \textbf{Qualitative evaluation (explanatory mechanisms).} Participants had to provide a textual description that explains the decision made for the interview variable. Optionally, participants could also submit a visual description to  enrich and improve clarity and explainability. Performance was evaluated in terms of the creativity of participants and the explanatory effectiveness of the descriptions. For this evaluation, we invited a set of experts in the fields of psychological behavior analysis, recruitment, machine learning and computer vision. 

Since the explainability component of the challenge requires qualitative evaluations and hence human effort, the scoring of participants was made based on a small subset of the videos. Specifically, subsets of videos from the validation and test sets were systematically selected to better represent the variability of the personality traits and invite-for-interview values in the entire dataset. The jury only evaluated a single validation and a single test phase submission per participant. A separate jury member served as a tiebreaker. At the end, the creativity criterion was judged globally, according to the evaluated clips, as well as an optional video that participants could submit to describe their method. Figure~\ref{fig:interface} shows an illustration of the  interface used by the jury for the qualitative evaluation phase.
\begin{figure}[htb!]
    \centering   
    \hspace{1cm}\includegraphics[scale=0.22]{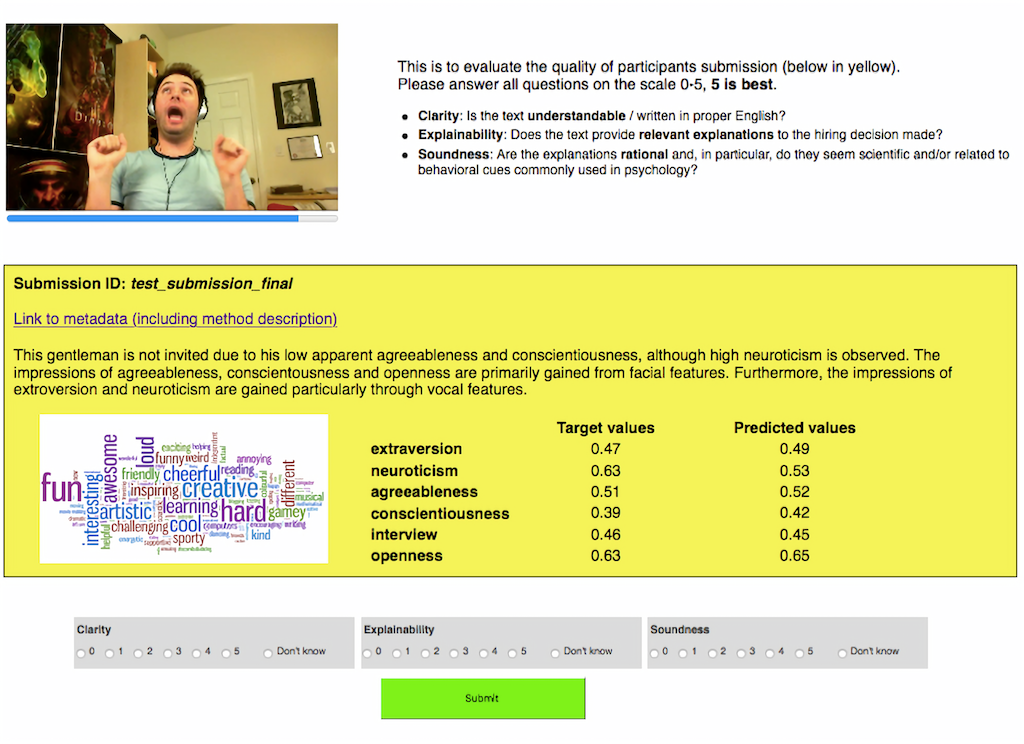} 
    \caption{Qualitative evaluation interface. The explainable interface of a submission is shown to the judge who had to evaluate it along the considered dimensions.}\label{fig:interface}
\end{figure}

For each evaluated clip, the evaluation criteria for the jury were: 
\begin{itemize}
\item \emph{Clarity:} Is the text understandable / written in proper English?

\item \emph{Explainability:} Does the text provide relevant explanations on the hiring decision made?

\item \emph{Soundness:} Are the explanations rational and, in particular, do they seem scientific and/or related to behavioral cues commonly used in psychology?



\end{itemize}

The following two criteria were evaluated globally, based on the evaluated clips and the optional submitted video.

\begin{itemize}
\item \emph{Model interpretability:} Are the explanations useful to understand the functioning of the predictive model?

\item \emph{Creativity:} How original / creative are the explanations?

\end{itemize}

\item \textbf{Coopetition evaluation (code sharing).} Participants were  evaluated by the usefulness of their shared code in the collaborative competition scheme. The coopetition scheme was implemented in the second stage of the challenge.








\end{itemize}


\subsection{Baselines}

We considered several baselines for solving the aforementioned tasks in different input modalities. Here, we briefly describe the baseline models and results (see \cite{Gucluturk2017,Gucluturk2017b} for more details). 
\edition{Please note that the factors underlying the predictions of the  (baseline) models for the quantitative phase were investigated in these two earlier publications \cite{Gucluturk2017,Gucluturk2017b}. Briefly, these two studies show that many factors including face features, gender, audio features and context have varying contributions to the predictions of different traits.}
\edition{Also note that we only describe the baseline adopted for the first stage for the quantitative stage of the challenge. For the qualitative evaluation we built a demo\footnote{\edition{\url{http://sergioescalera.com/wp-content/uploads/2016/12/TraitVideo.mp4}}} that was successfully presented at the demo session of NIPS 2016\footnote{\edition{\url{https://nips.cc/Conferences/2016/Schedule?showEvent=6314}}}. The purpose of the demo was to give an idea to participants on possibilities for designing their systems. }
\subsubsection{Language models: audio transcripts}

We evaluated two different language models, each on the same modality (transcriptions). Both of the models were a variation of the following (linearized) ridge regression model: $y = \embedding \left( \mathbf{ x } \right) \boldsymbol{ \beta } + \varepsilon$, 
%
where $y$ is the annotation, $\mathbf{ x }$ is the transcription, $\boldsymbol{ \beta }$ represents the parameters and $\varepsilon$ is the error term. This formulation describes a (nonlinear) embedding, followed by a (linear) fully-connected computation. Both models were trained by analytically minimizing the L2 penalized least squares loss function on the training set, and model selection was performed on the validation set.

\paragraph{Bag-of-words model.} This model uses an embedding that represents transcripts as 5000-dimensional vectors, i.e. the counts of the 5000 most frequent non-stopwords in the transcriptions.

\paragraph{Skip-thought vectors model.} This model uses an embedding that represents transcripts as 4800-dimensional mean skip-thought vectors \cite{Kiros2015skip} of the sentences in the transcriptions. A recurrent encoder-decoder neural network pretrained on the BookCorpus dataset \cite{Zhu2015aligning} was used for extracting the skip-thought vectors from the transcriptions.

\subsubsection{Sensory models: audio visual information processing}

We evaluated three different sensory models, each on a different modality (audio, visual, and audio visual, respectively). All models were a variation of the 18-layer deep residual neural network (ResNet18) in \cite{He2015}. As such, they comprised several convolutional layers followed by rectified linear units and batch normalization, and connected to one another with (convolutional or identity) shortcuts, as well as a final (linear) fully-connected layer preceded by global average pooling. The models were trained by minimizing the mean absolute error loss function iteratively with stochastic gradient descent (Adam \cite{Kingma2014}) on the training set, and model selection was performed on the validation set.

\paragraph{Audio model.} This model is a variant of the original ResNet18 model, in which $n \times n$ inputs, kernels, and strides are changed to $n^2 \times 1$ inputs, kernels, and strides \cite{Guclu2016brains}, as well as changing the size of the last layer to account for the different number of outputs. Prior to entering the model, the audio data were temporally preprocessed to 16kHz. The model was trained on random 3s crops of the audio data and tested on the entire audio data.

\paragraph{Visual model.} This model is a variant of the original ResNet18 model, in which the size of the last layer is changed to account for the different number of outputs. Prior to entering the model, the visual data are spatiotemporally preprocessed to $456 \times 256$ pixels and 25 frames per second. The model was trained on random $224 \times 224$ pixel single frame crops of the visual data and tested on the entire visual data.

\paragraph{Audiovisual model.} This model is obtained by a late fusion of the audio and visual models. The late fusion took place after the global average pooling layers of the models via concatenation of their latent features. The entire model was jointly trained from scratch.

\subsubsection{Language and sensory model}

\paragraph{Skip-thought vectors and audiovisual model.} This model is obtained by a late fusion of the pretrained skip-thought vectors and audiovisual models. The late fusion took place after the embedding layer of skip-thought vectors model and the global average pooling layer of the audiovisual model via concatenation of their latent features. Only the last layer was trained from scratch and the rest of the layers were fixed.

\subsubsection{Results}

The baseline models were used to predict the trait annotations as a function of the language and/or sensory data. 
Table~\ref{tab:baseline} shows the baseline results. The language models had the lowest overall performance with skip-thought vectors model performing better than the bag-of-word model. The performance of the sensory models were better than those of the language models with the audiovisual fusion model having the highest performance and the audio model having the lowest performance. Among all models, the language and sensory fusion model (skip-thought vectors and audiovisual fusion model) achieved the best performance. All prediction accuracies were significantly above the chance-level ($p < 0.05$, permutation test), and were consistently improved by fusing more modalities.

\begin{table}[]
\centering
\caption{Baseline results. Results are reported in terms of 1 - relative mean absolute error on the test set. $AGR$: Agreeableness; $CON$: Conscientiousness; $EXT$: Extroversion; $\overline{NEU}$: (non-)Neuroticism; $OPE$: Openness; $AVE$: average over trait results; $INT$: interview.}
\label{tab:baseline}
\begin{tabular}{@{}lccccccc@{}}
\toprule
Model & $AGR$ & $CON$ & $EXT$ & $\overline{NEU}$: & $OPE$ & $AVE$ & $INT$ \\ \midrule
\textbf{language} & \multicolumn{1}{l}{} & \multicolumn{1}{l}{} & \multicolumn{1}{l}{} & \multicolumn{1}{l}{} & \multicolumn{1}{l}{} & \multicolumn{1}{l}{} & \multicolumn{1}{l}{} \\
~bag-of-words & 0.8952 & 0.8786 & 0.8815 & 0.8794 & 0.8875 & 0.8844 & 0.8845 \\
~Skip-Thought Vec. & 0.8971 & 0.8819 & 0.8839 & 0.8827 & 0.8881 & 0.8867 & 0.8865 \\
\textbf{sensory} & \multicolumn{1}{l}{} & \multicolumn{1}{l}{} & \multicolumn{1}{l}{} & \multicolumn{1}{l}{} & \multicolumn{1}{l}{} & \multicolumn{1}{l}{} & \multicolumn{1}{l}{} \\
~audio & 0.9034 & 0.8966 & 0.8994 & 0.9000 & 0.9024 & 0.9004 & 0.9032 \\
~visual & 0.9059 & 0.9073 & 0.9019 & 0.8997 & 0.9045 & 0.9039 & 0.9076 \\
~Audio-Visual & 0.9102 & 0.9138 & 0.9107 & 0.9089 & 0.9111 & 0.9109 & 0.9159 \\
\textbf{language+sensory} & \multicolumn{1}{l}{} & \multicolumn{1}{l}{} & \multicolumn{1}{l}{} & \multicolumn{1}{l}{} & \multicolumn{1}{l}{} & \multicolumn{1}{l}{} & \multicolumn{1}{l}{} \\
~STV + AV & \textit{0.9112} & \textit{0.9152} & \textit{0.9112} & \textit{0.9104} & \textit{0.9111} & \textit{0.9118} & \textit{0.9162} \\ \bottomrule
\end{tabular}
\end{table}

\section{Two systems}
\label{sec:solutions_from_participants}
This section provides a detailed description of two systems that completed the second stage of the job candidate screening challenge.  \edition{Both methods have been briefly described previously, this section includes a comprehensive and detailed description and analysis of both techniques. }
\subsection{BU-NKU: Decision Trees for Modality Fusion and Explainable Machine Learning}
\label{section:bunku}
\begin{figure*}
	\centering
	\includegraphics[width=0.95\textwidth]{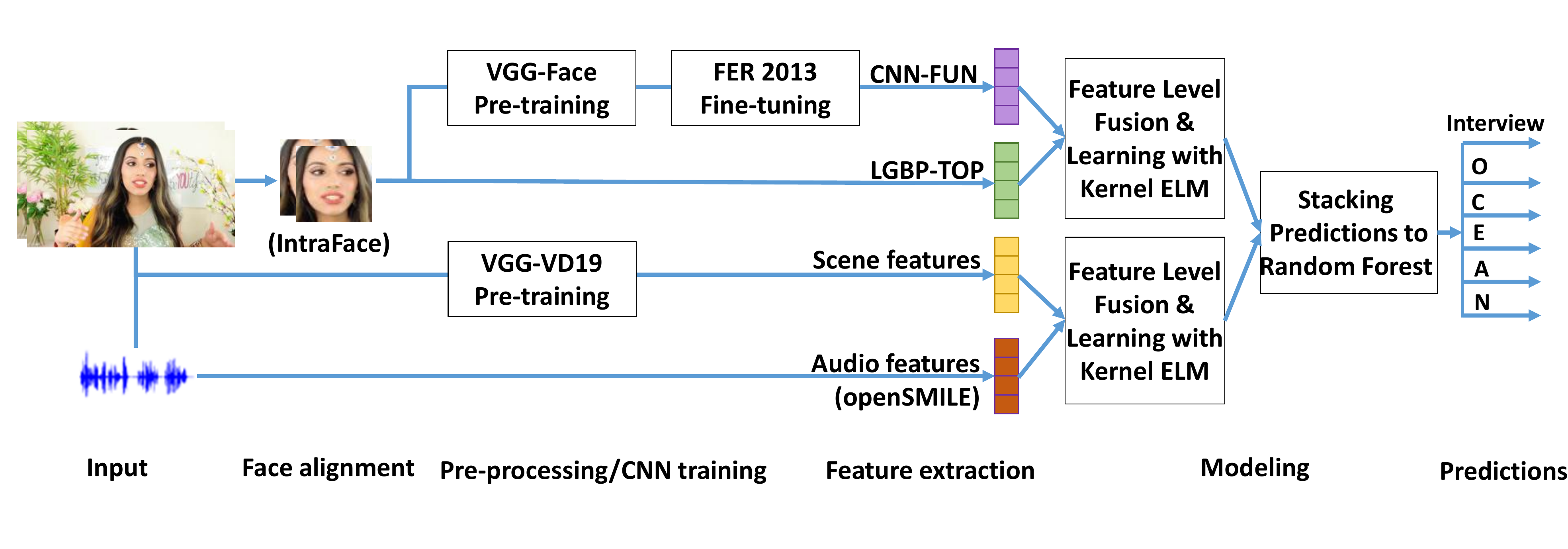}
	\caption{Flowchart of the BU-NKU system.}
	\label{fig:flowchart}
\end{figure*}

The BU-NKU 
system is based on audio, video, and scene features. A similar pipeline was used in the system that won the ChaLearn First Impression Challenge at ICPR 2016~\cite{Gurpinar_2016_ICPR_Workshops}. The main difference is that here, the face, scene, and audio modalities are first combined at feature level, followed by stacking the predictions of sub-systems to an ensemble of decision trees~\cite{Kaya_2017_CVPR_Workshops}. The flow of this system is illustrated in Figure~\ref{fig:flowchart}.
The qualitative stage inputs the final predictions from the system proposed for the quantitative stage, discretizes them using the training set mean scores of each target dimension, then maps the binarized (low/high) personality traits to the binarized (invite/do not invite) interview variable via a decision tree (DT). DT is employed to allow visualization and ease of interpretation for the model, hence allow explainability for the decision. Finally the DT is traced to generate a verbal explanation. A wider but brief summary of the components used in this system is provided in the subsequent subsections. 
\subsubsection{Quantitative System}
\label{section:bu-nku-quan}
Facial features are extracted over an entire video segment and summarized by functionals. Scene features, however, are extracted from the first image of each video only. The assumption is that videos do not stretch over multiple shots. 
\paragraph{Face Features}
Faces are detected on all frames of the video input. The Supervised Descent Method (SDM) is used for face registration, which gives 49 landmarks on each detected face~\cite{XiongCVPR2013}. The roll angle is estimated from the eye corners to rotate the image accordingly. Then a margin of 20\% of the interocular distance around the outer landmarks is added to crop the facial image. Each image is resized to $64\times 64$ pixels.

After aligning the faces, image-level deep features are extracted from a convolutional neural network trained for facial emotion recognition. To prepare this feature extractor, the system starts with the pre-trained VGG-Face network~\cite{Parkhi15}, which is optimized for the face recognition task on a very large set of faces. Then this network is fine-tuned for emotion~\cite{kaya2017imavis}, using more than 30K training images of the FER-2013 dataset~\cite{goodfellow2013challenges}. 
The final trained network has a 37-layer architecture (involving 16 convolution layers and 5 pooling layers). The response of the 33\edition{rd} layer is used, which is the lowest-level 4\,096-dimensional descriptor. 

After extracting frame-level features from each aligned face, videos are summarized by computing functional statistics of each dimension over time. The functionals include mean, standard deviation, offset, slope, and curvature. Offset and slope are calculated from the first order polynomial fit to each feature contour, while curvature is the leading coefficient of the second order polynomial. 

In the BU-NKU approach, deep facial features are combined with the Local Gabor Binary Patterns from Three Orthogonal Planes (LGBP-TOP) video descriptor, shown to be effective in emotion recognition~\cite{almaev2013local,kaya2017imavis}. 
It is extracted by applying 18 Gabor filters on aligned facial images with varying orientation and scale parameters. The resulting feature dimensionality is 50\,112.
\paragraph{Scene Features}
In order to use ambient information in the images, a set of features is extracted using the VGG-VD-19 network~\cite{simonyan2014very}, which is trained for an object recognition task on the ILSVRC 2012 dataset. Similar to face features, a 4\,096-dimensional representation from the 39$^{th}$ layer of the 43-layer architecture is used. This gives a description of the overall image that contains both face and scene. The effectiveness of scene features for predicting Big Five traits is shown in~\cite{gurpinar2016eccv,Gurpinar_2016_ICPR_Workshops}. For Job Candidate Screening task, these features contribute to the final decision both directly and indirectly over the personality trait predictions.
\paragraph{Acoustic Features}
The open-source openSMILE tool~\cite{eyben2010opensmile} is popularly used to extract acoustic features in a number of international paralinguistic and multi-modal challenges. The idea is to obtain a large pool of potentially relevant features by passing an extensive set of summarizing functionals on the low level descriptor contours (e.\,g. Mel Frequency Cepstral Coefficients, pitch, energy and their first/second order temporal derivatives). The BU-NKU approach uses the toolbox with a standard feature configuration that served as the challenge baseline sets in INTERSPEECH 2013 Computational Paralinguistics Challenge~\cite{schuller2013interspeech}. This configuration was found to be the most effective acoustic feature set among others for personality trait recognition~\cite{Gurpinar_2016_ICPR_Workshops}. 

\paragraph{Model Learning} In order to model personality traits from audio-visual features, kernel extreme learning machines (ELM) were used, due to the learning speed and accuracy of the algorithm. Initially, ELM is proposed as a fast learning method for Single Hidden Layer Feedforward Networks (SLFN): an alternative to back-propagation~\cite{huang2004extreme}. To increase the robustness and the generalization capability of ELM, a regularization coefficient is included in the optimization procedure. 
\paragraph{Score Fusion} The predictions of the multi-modal ELM models are stacked to a Random Forest (RF), which is an ensemble of decision trees (DT) grown with a random subset of instances (sampled with replacement) and a random subset of features~\cite{breiman2001random}. Sampling with replacement leaves approximately one third of the training set instances \textit{out-of-bag}, which are used to cross-validate the models and optimize the hyper-parameters at the training stage. This is an important aspect of the method regarding the challenge conditions, as cross validation gives an unbiased estimate of the expected value of prediction error~\cite{breiman1996heuristics}.

The validation set performances of individual features, as well as their feature-, score- and multi-level fusion alternatives are shown in Table~\ref{tab:bu_nku_feature_performances}. Here, System 0 corresponds to the top entry in the ICPR 2016 Challenge~\cite{Gurpinar_2016_ICPR_Workshops}, which uses the same set of features and fuses scores with linear weights. For the weighted score fusion, the weights are searched in the [0,1] range with steps of 0.05. Systems 1 to 6 are sub-components of the proposed system, namely System 8, whereas System 7 is a score fusion alternative that uses linear weights instead of a Random Forest. Systems 1 and 2 are trained with facial features as explained before: VGGFER33 is $33^{rd}$ layer output of FER fine-tuned VGG CNN and LGBPTOP is also extracted from face. These two facial features are combined in the proposed framework, and their feature-level fusion performance is shown as System 5. Similarly, Systems 3 (scene sub-system) and 4 (audio sub-system) are combined at feature level as System 6.  

In general, fusion scores are observed to benefit from complementary information of individual sub-systems. Moreover, we see that fusion of face features improve over their individual performance. Similarly, the feature level fusion of audio and scene sub-systems is observed to benefit from complementarity. The final score fusion with RF outperforms weighted fusion in all but one dimension (agreeableness), where the performances are equal.
\begin{table*}[htbp]
	\centering
	\caption{Validation set performance of the BU-NKU system and its sub-systems, using the performance measure of the challenge (1-relative mean abs error). FF: Feature-level fusion, WF: Weighted score-level fusion, RF: Random Forest based score-level fusion. INTER: Interview invite variable. $AGRE$: Agreeableness. $CONS$: Conscientiousness. $EXTR$: Extroversion. $\overline{NEUR}$: (non-)Neuroticism. $OPEN$: Openness to experience.}
	\begin{tabular}{clcccccc}
				\hline
		\textbf{\#} & \multicolumn{1}{c}{\textbf{System}} & INTER & $AGRE$ & $CONS$ & $EXTR$ & $\overline{NEUR}$ & $OPEN$ \\		\hline
		0     & ICPR 2016 Winner & N/A   & 0.9143 & 0.9141 & 0.9186 & 0.9123 & 0.9141\\
		1     & Face: VGGFER33 & 0.9095 & 0.9119 & 0.9046 & 0.9135 & 0.9056 & 0.9090  \\
		2     & Face: LGBPTOP & 0.9112 & 0.9119 & 0.9085 & 0.9130 & 0.9085 & 0.9103  \\
		3     & Scene: VD\_19 & 0.8895 & 0.8954 & 0.8924 & 0.8863 & 0.8843 & 0.8942  \\
		4     & Audio: OS\_IS13 & 0.8999 & 0.9065 & 0.8919 & 0.8980 & 0.8991 & 0.9022  \\
		5     & FF(Sys1, Sys2) & 0.9156 & 0.9144 & 0.9125 & 0.9185 & 0.9124 & 0.9134  \\
		6     & FF(Sys3, Sys4) & 0.9061 & 0.9091 & 0.9027 & 0.9013 & 0.9033 & 0.9068  \\
		7     & WF(Sys5, Sys6) & 0.9172 & \textbf{0.9161} & 0.9138 & 0.9192 & 0.9141 & 0.9155  \\
		8     & RF(Sys5, Sys6) & \textbf{0.9198} & \textbf{0.9161} & \textbf{0.9166} & \textbf{0.9206} & \textbf{0.9149} & \textbf{0.9169}  \\
				\hline
	\end{tabular}%
	\label{tab:bu_nku_feature_performances}%
\end{table*}%

Based on the validation set results, the best fusion system (System 8 in Table~\ref{tab:bu_nku_feature_performances}) is obtained by stacking the predictions from Face feature-fusion (FF) model (System 5) with the Audio-Scene FF model (System 6). This fusion system renders a test set performance of 0.9209 for the interview variable, ranking the first and beating the challenge baseline score.
\subsubsection{Qualitative System}
For the qualitative stage, the final predictions from the RF model are binarized by thresholding each score with its corresponding training set mean value. The binarized predicted OCEAN scores are mapped to the binarized ground truth interview variable using a decision tree (DT) classifier. The use of a DT is motivated by the fact that the resulting model is self-explanatory and can be converted into an explicit recommender algorithm using ``if-then" rules. The proposed approach for decision explanation uses the trace of each decision from the root of the tree to the leaf. The verbal explanations are finally accompanied with the aligned image from the first face-detected frame and the bar graphs of corresponding mean normalized scores.

\begin{figure*}
	\centering
	\includegraphics[width=.8\textwidth]{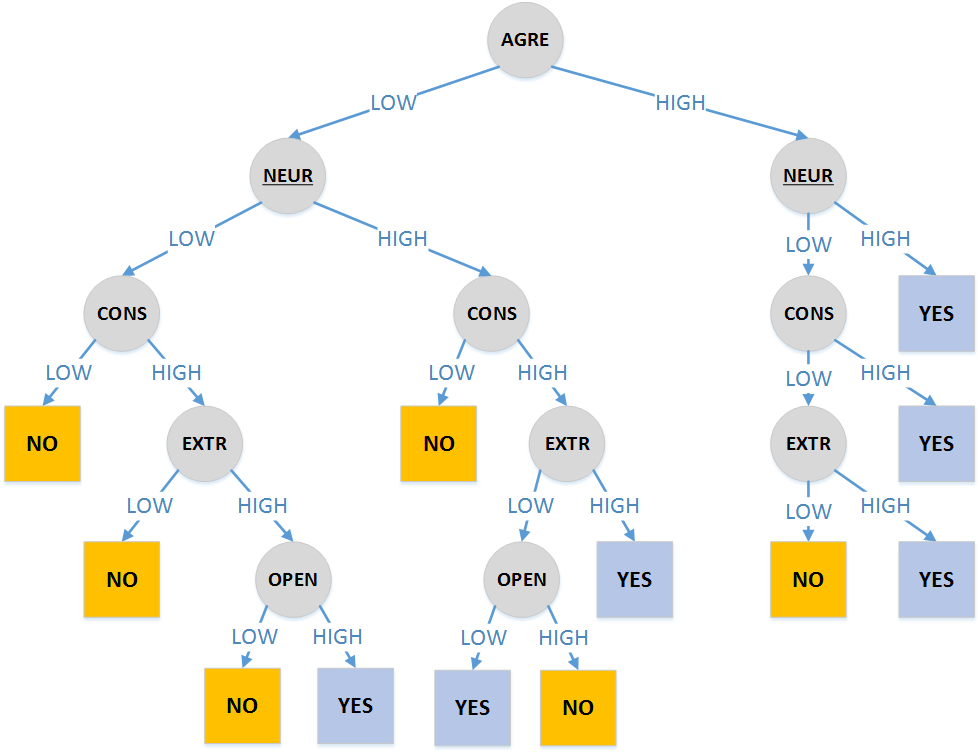}
	\caption{Illustration of the decision tree for job interview invitation. \underline{NEUR} denotes (non-) Neuroticism. Leaves denote a positive or a negative invitation response.} 
	\label{fig:dt}
\end{figure*}

\begin{figure*}
	\centering
\begin{tabular}{p{0.45\textwidth} p{0.45\textwidth}}
\begin{center}
\includegraphics[bb= 0 0 355 200,width=0.3\textwidth]{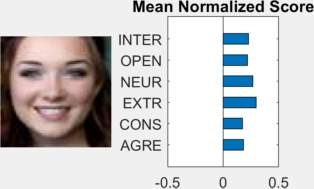}	
\end{center}
&
 \begin{center}
 \includegraphics[bb= 0 0 355 200,width=0.3\textwidth]{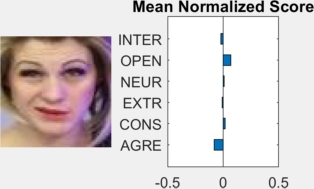} 
\end{center}
   \\
\emph{This lady is invited for an interview due to her high apparent agreeableness and non-neuroticism impression. The impressions of agreeableness, conscientiousness, extroversion, non-neuroticism and openness are primarily gained from facial features.} & 
\emph{This lady is not invited for an interview due to her low apparent  agreeableness and extraversion impressions, although predicted scores for non-neuroticism, conscientiousness and openness were high. It is likely that this trait combination (with low agreeableness, low extraversion, and high openness scores) does not leave a genuine impression for job candidacy. The impressions of agreeableness, extraversion, non-neuroticism and openness are primarily gained from facial features. Furthermore, the impression of conscientousness is predominantly modulated by voice.} 
\\
\end{tabular}
	\caption{Sample verbal and visual explanations from qualitative stage for the BU-NKU entry.}
	\label{fig:samples}
\end{figure*}

The DT trained on the predicted OCEAN dimensions gives a classification accuracy of 94.2\% for binarized interview variable. The illustration of the trained DT is given in Figure~\ref{fig:dt}. 
The model is intuitive as higher scores of traits generally increase the chance of interview invitation. As can be seen from the figure, the DT ranks relevance of the predicted Big Five traits from highest (Agreeableness) to lowest (Openness to Experience) with respect to information gain between corresponding trait and the interview variable. The second most important trait for job interview invitation is Neuroticism, which is followed by Conscientiousness and Extroversion. The high/low scores of these top four traits are correlated with target variable and are observed to be consistent throughout the DT. If the Openness score is high, then having a high score in any of the variables (non-)Neuroticism, Conscientiousness or Extroversion suffices for invitation. Chances of invitation decrease if Agreeableness is low: only three out of eight leaf nodes are ``YES" in this branch. In two of these cases, one has to have high scores in three out of four remaining traits. Figure~\ref{fig:samples} illustrates automatically generated verbal and visual explanations for this stage. 

\subsection{TUD: Layered Linear Regression and Weight Analysis}
This section describes the TUD approach for the second stage of the job candidate screening challenge. 
This system was particularly designed to give assistance to a human assessor. The proposed model employs features that can easily be described in natural language, with a linear (PCA) transformation to reduce dimensionality, and simple linear regression models for predicting scores, such that scores can be traced back to and justified with the underlying features. While state-of-the-art automatic solutions rarely use hand-crafted features and models of such simplicity, there are clear gains in explainability. As demonstrated within the ChaLearn benchmarking campaign, this model did not obtain the strongest quantitative results, but the human-readable descriptions it generated were well appreciated by human judges.

The model considers two modalities, visual and textual, for extracting features. In the visual modality, it considers features capturing facial movement and expression, as they are one of the best indicators for personality~\cite{Naumann2009,Borkenau2009}. However, considering findings in organizational psychology, personality traits are not the only (and neither the strongest) predictors for job suitability. In fact, GMA (General Mental Ability) tests, such as intelligence tests, have the highest validity at the lowest application cost~\cite{Schmidt1998,Cook2009}. While there is no formal GMA assessments for subjects in the dataset, language use of the vlogger may indirectly reveal GMA characteristics, such as the use of difficult words. Consequently, textual features, including speaking density, as well as linguistic sophistication, were considered for this approach.

\subsubsection{Visual Features}

For the visual representation, the system was not built to focus on the video in general, but particularly on facial expression and movement. 
OpenFace tools were used to segment only the face from each video~\cite{amos2016openface}, standardizing the segmented facial video to 112$\times$112 pixels.
OpenFace is an open source toolkit which does not only segment faces, but offers a feature extraction library that can extract and characterize facial movements and gaze~\cite{Baltrusaitis2015}.  OpenFace is able to recognize a subset of individual Action Units (AU) that construct facial expressions encoded in Facial Action Code System (FACS) as shown in Table \ref{tab:actionunit}~\cite{Ekman1978a,Ekman2005}. These AUs then can be described in two ways: in terms of presence (indicating whether a certain AU is detected in a given time frame) and intensity (indicating how intense an AU is at a given time frame).

\begin{table}[h]
	\caption{Action Units that are recognized by OpenFace.}
	\label{tab:actionunit}
	\begin{center}
		\begin{tabular}{llll}
			\hline
			Action Unit & Description & Action Unit & Description \\
			\hline
			AU1 & Inner Brow Raiser & AU14 & Dimpler \\
			AU2 & Outer Brow Raiser & AU15 & Lip Corner Depressor \\
			AU4 & Brow Lowerer & AU17 & Chin Raiser \\
			AU5 & Upper Lid Raiser & AU20 & Lip stretcher \\
			AU6 & Cheek Raiser & AU23 & Lip Tightener \\
			AU7 & Lid Tightener &AU25 & Lips part \\	
			AU9 & Nose Wrinkler &AU26 & Jaw Drop \\
			AU10 & Upper Lip Raiser &AU28 & Lip Suck \\
			AU12 & Lip Corner Puller &AU45 & Blink \\
			\hline
		\end{tabular}
	\end{center}
\end{table}

For each of these AUs, three features were constructed for input to the system. First, the percentage of time frames is computed, during which the AU was visible in a video. Second, the maximum intensity of the AU in the video was stored. Lastly, the mean intensity of the AU over the video was also recorded. These three features per AU add up to 52 features in total for the OpenFace representation.

\begin{figure}[h]
	\centering
	\includegraphics[width=0.8\linewidth]{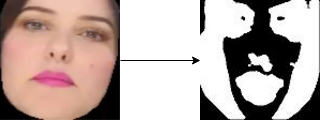}
	\caption{wMEI for face segmented video.}
	\label{fig:meiface}
\end{figure}

The resulting segmented video is also used for another video representation. In order to capture overall movement of the vlogger's face, a Weighted Motion Energy Image (wMEI) is constructed from the resulting face segmented video. MEI is a grayscale image that shows how much movement happens on each pixel throughout video, with white indicating a lot of movement and black indicating less movement~\cite{Bobick2001}. wMEI was proposed in the work of Biel et al.~\cite{Biel2011} as a normalized version of MEI, by dividing each pixel values with the maximum pixel value. 
The method proposed by TUD is inspired by the aforementioned work with improvement on background noise reduction. In~\cite{Biel2011}, the whole video frame is used as an input to compute wMEI, which makes background movement contribute to the overall wMEI measurements. Thus, there are cases in which the resulting wMEI is all white due to background or camera movements, rather than movement of a human subject. For example, this happens when the vlogger recorded the video in a public space or while on the road. Using the face segmented video instead of a whole video frame, the involvement of background is minimized to get a better representation of the subject's true movement, as can be seen in Figure~\ref{fig:meiface}. In order to create wMEI, the base face image of each video is obtained and the overall movement for each pixel is computed over video frames. For each wMEI, three statistical features (mean, median, and entropy) are extracted to constitute a MEI representation.

The ChaLearn dataset has been carefully prepared so that only one unique foreground person faces the camera in the video. However, the current OpenFace implementation has limitations when the video still contains other visual sources with faces, such as posters or music covers in the background. While the situation is rare, it was occasionally noticed that a poster was detected and segmented as `main face' rather than the subject's actual face. For such misdetections, no movement will be detected at all, so these outliers are easily captured by the system.

\subsubsection{Textual Features}

Textual features are generated by using transcripts that were provided as the extension of the ChaLearn dataset. For a handful of videos, transcript data was missing; those videos were manually annotated, such that all videos have a transcript, with exception of one video in which the person speaks in sign language. 

As reported in the literature~\cite{Schmidt1998,Cook2009} and confirmed in private discussions with organizational psychologists, assessment of GMA (intelligence, cognitive ability) is important for many hiring decisions. While this information is not reflected in personality traits, language usage of the subjects may possibly reveal some related information. To assess that, several Readability indices were used with the transcripts. This was done by using open source implementations of various readability measures in the NLTK-contrib package of the Natural Language Toolkit (NLTK). More specifically, eight measures were selected as features for the Readability representation: ARI~\cite{Smith1967}, Flesch Reading Ease~\cite{Flesch1948}, Flesch-Kincaid Grade Level~\cite{Kincaid1975}, Gunning Fog Index~\cite{gunning1952technique}, SMOG Index~\cite{McLaughlin1969}, Coleman Liau Index~\cite{Coleman1975}, LIX, and RIX~\cite{LIX/RIX}. While these measures are originally developed for written text (and ordinarily may need longer textual input than a few sentences in a transcript), they do reflect complexity in language usage. In addition, two simple statistical features were used for an overall Text representation: total word count in the transcript, and the amount of unique words within the transcript, respectively. 

\subsubsection{Quantitative System}

\begin{figure}[htb]
	\centering
	\includegraphics[width=.8\linewidth]{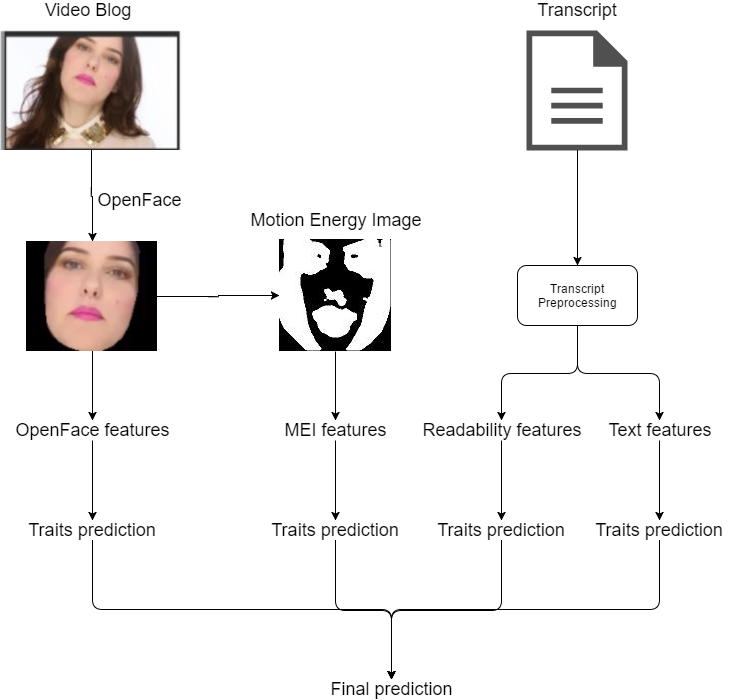}
	\caption{Overall system diagram for the TUD system.}
	\label{fig:diagram}
\end{figure}

The building blocks of the TUD predictive model encompass four feature representations; OpenFace, MEI, Readability, and Text. Employing the 6000 training set videos, for each representation, a separate model was trained to predict personality traits and interview scores. For a final prediction score, late fusion is used and the predictions made by the four different models were averaged. A diagram of the proposed system can be seen in Figure \ref{fig:diagram}.

As the goal of the system is to trace back the prediction scores to each underlying feature, linear models were selected. Linear regression is a commonly used model in the social sciences literature. Principal Component Analysis (PCA) is used to reduce dimensionality, retaining 90\% variance. The resulting transformed features are used as input for a simple linear regression model to predict the scores.

The linear regression coefficients indicate for each PCA dimension whether it contributes positively or negatively to the prediction. Furthermore, considering the PCA transformation matrix, the strength of the contribution for original features can be determined.

\begin{table}[h]
	\caption{Accuracy (1 - Relative Mean Absolute Error) comparison between the TUD system and the lowest and highest accuracy for each prediction category in the ChaLearn CVPR 2017 Quantitative Challenge.}
	\label{tab:accuracy}
	\begin{center}
		\begin{tabular}{lccc}
			\hline
			Categories & TUD System & Lowest & Highest \\
			\hline
			Interview & 0.8877 & 0.8721 & 0.9209 \\
			$Agreeableness$ & 0.8968 & 0.8910 & 0.9137 \\
			$Conscientiousness$ & 0.8800 & 	0.8659 & 	0.9197 \\
			$Extroversion$ & 0.8870 & 0.8788 & 0.9212 \\
			$\overline{Neuroticism}$ & 0.8848 & 0.8632 & 0.9146 \\
			$Openness$ & 0.8903 & 	0.8747 & 0.9170 \\
			\hline
		\end{tabular}
	\end{center}
\end{table}

Table~\ref{tab:accuracy} shows the overall quantitative accuracy of the TUD system on the 2000 videos in the benchmark training set, for each of the Big Five personality traits and the interview invitation assessment. For each predicted class, scores are compared to the lowest and highest scores (from all of the participants) in the ChaLearn CVPR 2017 Quantitative Challenge.

While the system did not achieve the top scores, it was parsimonious in its use of computational resources, and the linear models allowed easier explainability. This is clearly a trade-off in such systems, as more parameters in the model and increased complexity makes interpretation more difficult.

\subsubsection{Qualitative System}
In the Qualitative phase of the ChaLearn CVPR 2017 Challenge, the goal was to explain predictions with a human-understandable text. The TUD system implements a simple text description generator, with the following justifications:
\begin{itemize}
	\item 
    Each of the visual and textual features were picked to be explainable in natural language to non-technical users. However, no formal proof was given that the features are fully valid predictors of personality traits or interviewability. While the model gives indicators on the strongest linear coefficients, the assessments it was trained on are made by external observers (crowdsourcing workers), which poses a very different situation from the assessment settings in the formal psychology studies, as reported in~\cite{Schmidt1998}. Therefore, these features do not constitute a comprehensive panel of ``good'' features, despite their good predictive power.
	\item It may be possible to aggregate feature observations to higher-level descriptions (in particular, regarding AU detections, as combinations of AUs may indicate higher-level emotional expressions), but as this would increase the complexity of the model, only a basic explanation using individual low-level features was kept.
	\item As the feature measurements did not formally get tested (yet) in terms of psychometric validity, it is debatable to consider feature measurements and predicted scores as absolute indicators of interviewability. However, for each person, it was indicated whether the person scores ``unusually'' with respect to a larger population of ``representative subjects'' (formed by the vloggers represented in the 6000-video training set). Therefore, for each feature measurement, the system reports the typical range of the features and the percentile of the subject, compared to scores in the training set.
	\item Finally, to reflect major indicators from the linear model in the description, for each representation (OpenFace, MEI, Readability, Text) the two largest linear regression coefficients that were picked. For PCA dimensions corresponding to these coefficients, the features contributing most strongly to these dimensions were traced back, and their sign is checked. For these features, a short notice is added to the description, expressing how the feature commonly affects the final scoring (e.g.\ `In our model, a higher score on this feature typically leads to a higher overall assessment score') for a positive linear contribution.
\end{itemize}

As a result, for each video in the validation and test set, a fairly long, but consistent textual description was generated. An example fragment of the description is given in Figure~\ref{fig:description}.

\begin{figure*}[h]
	\begin{allintypewriter}
		*******************
		
		* USE OF LANGUAGE *
		
		*******************
		\vspace{4mm}\\
		Here is the report on the person's language use:
		\vspace{2mm}\\
		
		** FEATURES OBTAINED FROM SIMPLE TEXT ANALYSIS **
		
		Cognitive capability may be important for the job. I looked at a few very simple text statistics first.
		\vspace{2mm} \\
		*** Amount of spoken words ***
		
		This feature typically ranges between 0.000000 and 90.000000. The score for this video is 47.000000 (percentile: 62).
		
		In our model, a higher score on this feature typically leads to a higher overall assessment score.
	\end{allintypewriter}
	
	\caption{Example description fragment for the TUD system.}
	\label{fig:description}
\end{figure*}

\subsection{Challenge results}
\subsubsection{Stage 1 results: Recognizing first impressions}
\label{subsec:stage1results}

For the first stage, 72 participants registered for the challenge. 
Four valid submissions were considered for the prizes as summarized  in Table~\ref{tab:results}. The leading evaluation measure is the Recall of the Invite-for-interview variable (see Equation~\ref{eq:acc}), although we also show results for personality traits.
\begin{table}[h!tb]
   \caption{Results of the first stage of the job screening coopetition. * Leading evaluation measure.  }
	\label{tab:results}
   \begin{center}
   	\resizebox{\columnwidth}{!}{%
       \begin{tabular}{cccccccc} 
       \hline
\textbf{Rank}&\textbf{Team}& INTER & $AGRE$ & $CONS$ & $EXTR$ & $\overline{NEUR}$ & $OPEN$\\\hline
1&BU-NKU~\cite{Kaya_2017_CVPR_Workshops}&0.9209 (1)&0.9137 (1)&0.9197 (1)&0.9212 (1)&0.9146 (1)&0.9170 (1)\\ 
-&Baseline~\cite{Gucluturk2017}&0.9162 (2)&0.9112 (2)&0.9152 (2)&0.9112 (3)&0.9103 (2)&0.9111 (2)\\
2&PML~\cite{Bekhouche:CVPRW2017}&0.9157 (3)&0.9103 (3)&0.9137 (3)&0.9155 (2)&0.9082 (3)&0.9100 (3)\\ 
3&ROCHCI&0.9018 (4)&0.9032 (4)&0.8949 (4)&0.9026 (4)&0.9011 (4)&0.9047 (4)\\
4&FDMB&0.8721 (5)&0.8910 (5)&0.8659 (5)&0.8788 (5)&0.8632 (5)&0.8747 (5)\\\hline
\edition{-} & \edition{Prior} & \edition{0.8817} &\edition{0.8935} & \edition{0.8745}  & \edition{0.8780} & \edition{0.8773} & \edition{0.8833} \\ \hline
       \end{tabular}
       }
   \end{center}
\end{table}

\begin{figure}[thpb]
\centering
\subfigure{\includegraphics[height=3.0cm]{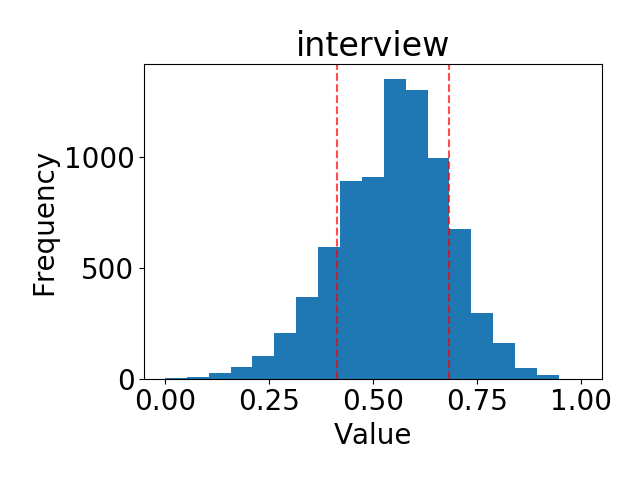}}\hspace{-0.2cm}
\subfigure{\includegraphics[height=3.0cm]{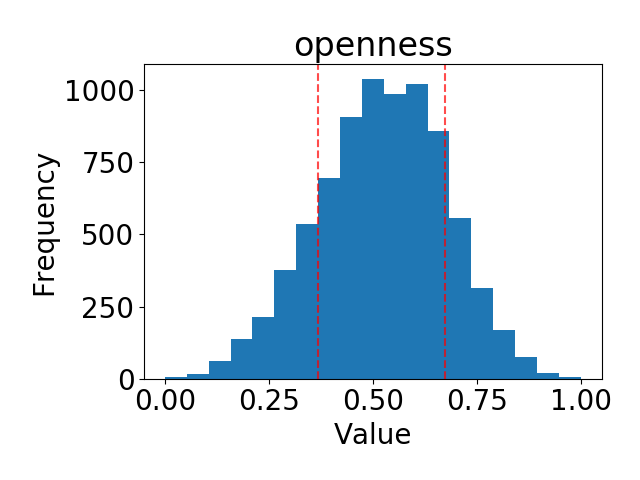}}\hspace{-0.2cm}
\subfigure{\includegraphics[height=3.0cm]{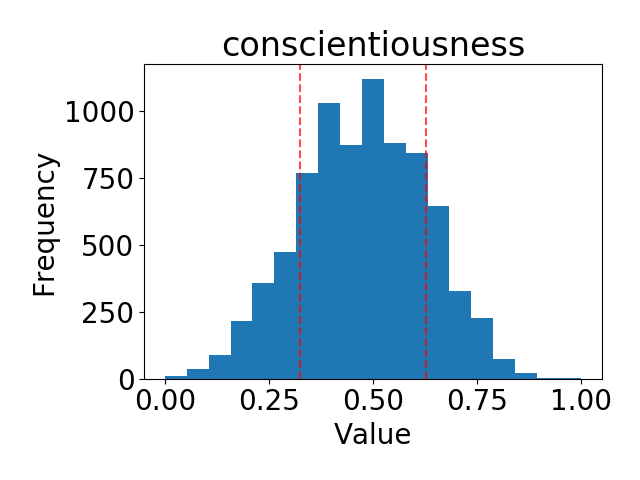}}\hspace{-0.2cm}
\subfigure{\includegraphics[height=3.0cm]{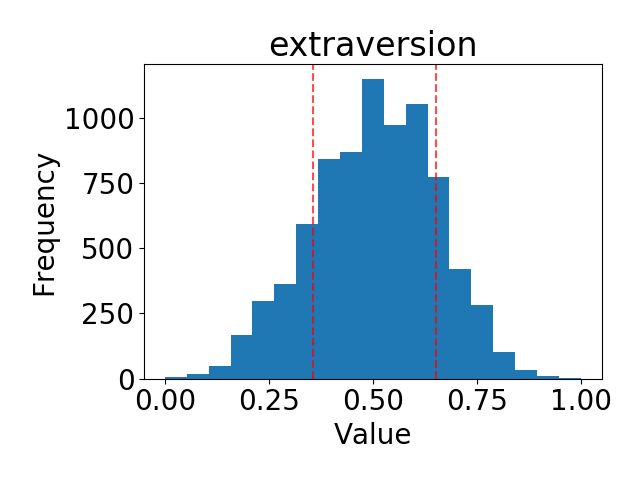}}\hspace{-0.2cm}
\subfigure{\includegraphics[height=3.0cm]{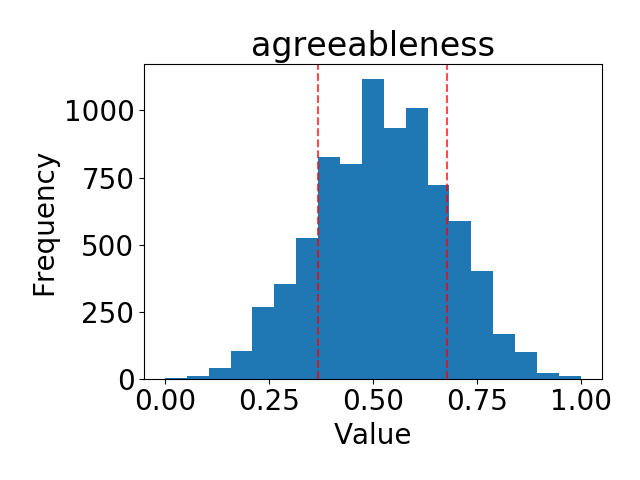}}\hspace{-0.2cm}
\subfigure{\includegraphics[height=3.0cm]{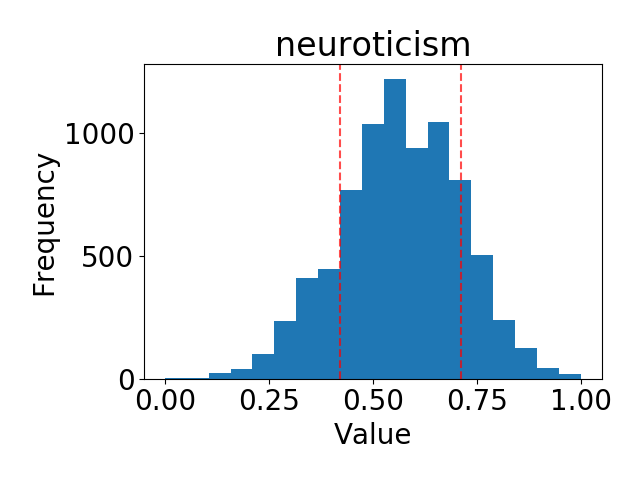}}\hspace{-0.2cm}
\caption{\edition{Distribution of the labels for each trait on the train set. Dashed lines show the deviation ($\sigma$) around the mean.}}
\label{hist:datasettrain}
\end{figure}

\begin{figure}[thpb]
\centering
\subfigure{\includegraphics[height=3.0cm]{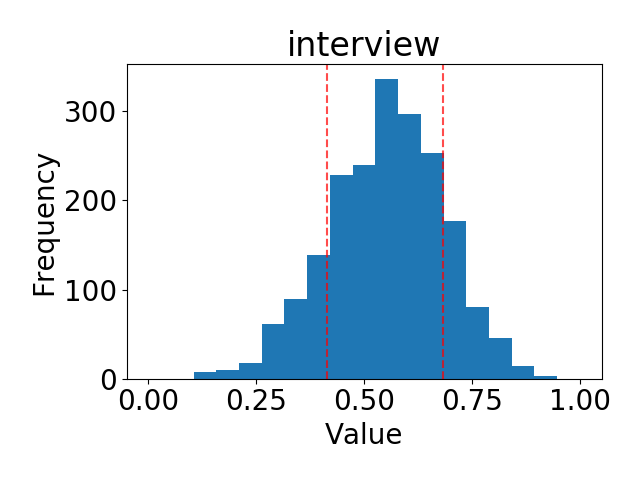}}\hspace{-0.2cm}
\subfigure{\includegraphics[height=3.0cm]{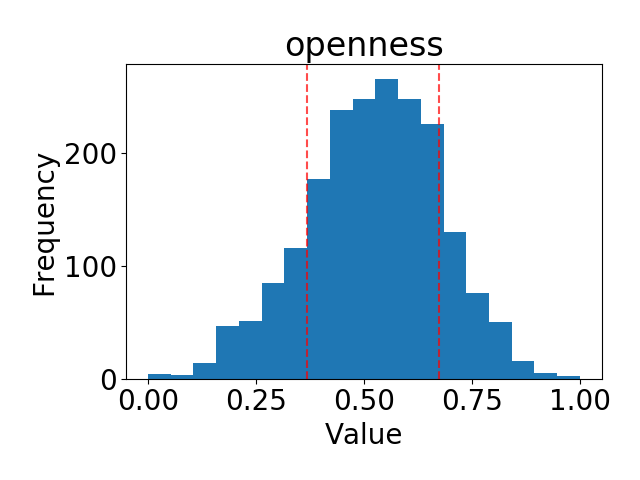}}\hspace{-0.2cm}
\subfigure{\includegraphics[height=3.0cm]{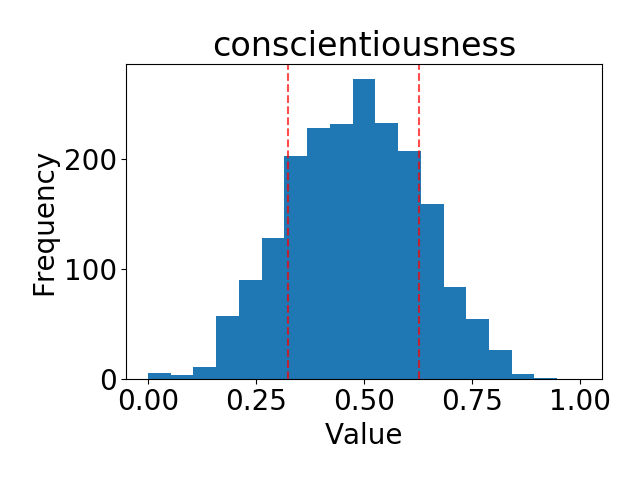}}\hspace{-0.2cm}
\subfigure{\includegraphics[height=3.0cm]{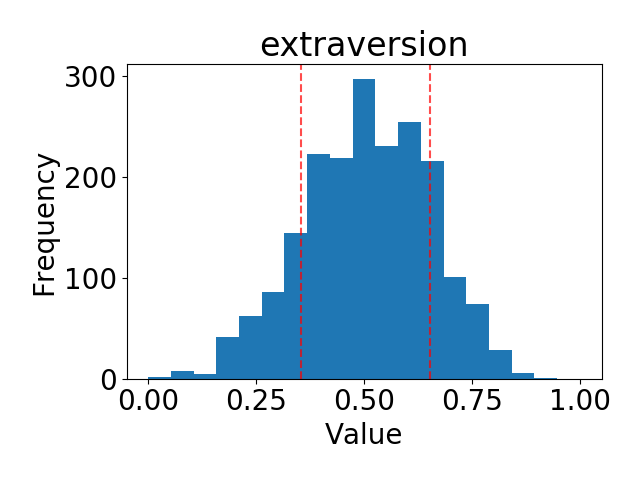}}\hspace{-0.2cm}
\subfigure{\includegraphics[height=3.0cm]{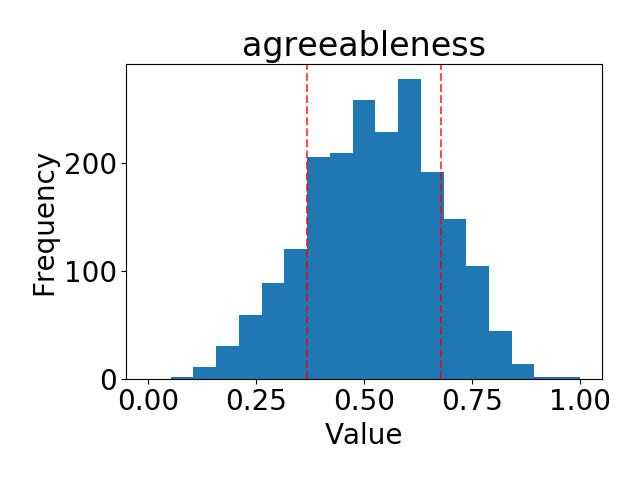}}\hspace{-0.2cm}
\subfigure{\includegraphics[height=3.0cm]{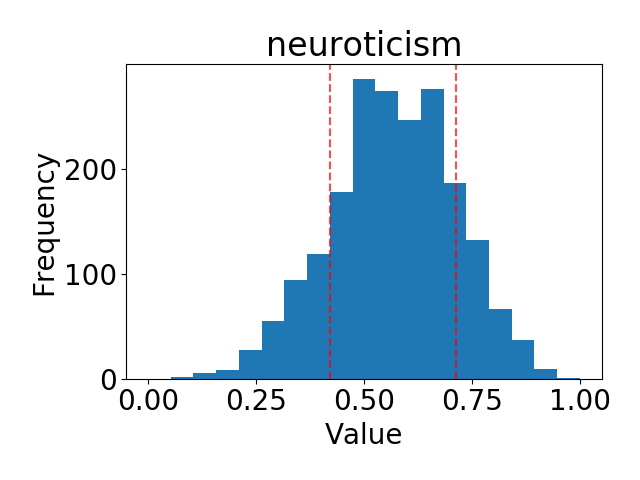}}\hspace{-0.2cm}
\caption{\edition{Distribution of the labels for each trait on the test set. Dashed lines show the deviation ($\sigma$) around the mean.}}
\label{hist:datasettest}
\end{figure}

Interestingly, only one out of the four valid submissions outperformed the baseline method described in the previous section. 
\edition{In addition to the baseline, we include in Table~\ref{tab:results} for reference the performance obtained if one would predict the average per-trait value for test instances (Prior). This simple prior would outperform the FDMB method ranked $4^{th}$. Although this was somewhat concerning (e.g., because a potential weakness of the data set), we found this was due to the similarity on the distribution of per-trait values between training and test partitions. This is illustrated in Figures~\ref{hist:datasettrain} and~\ref{hist:datasettest}.} 

\edition{The figures illustrate that by predicting the average value is a safe guess, however, this baseline will not work for predicting tail values (Section~\ref{sec:experimental_quant_cls}  reports additional results approaching the problem as one of classification). 
Nevertheless, this result suggests that additional metrics that account for tail prediction could be considered for a further analysis of methods performance.    
}

The performance of the top three methodologies was quite similar, however the methodologies were not. In the following, we provide a short description of the different methods.  

\begin{itemize}
\item \textbf{BU-NKU.} This approach was detailed in Section~\ref{section:bunku}, see~\cite{Kaya_2017_CVPR_Workshops} for further details.

\item \textbf{PML.~\cite{Bekhouche:CVPRW2017}} Adopted a purely visual approach based on multi-level appearance. After face detection and normalization,  Local Phase Quantization (LPQ) and Binarized Statistical Image Features (BSIF) descriptors were extracted at different scales of each frame using a grid. Feature vectors from each region and each resolution were concatenated, the representation for a video was obtained by averaging the per-frame descriptors. For prediction, the authors resorted in a stacking formulation: personality traits are predicted with  Support Vector regression (SVR), the outputs of these models are used as inputs for the final decision model, which, using Gaussian processes, estimates the invite for interview variable.  

\item \textbf{ROCHCI.} Extracted a set of predefined multi-modal features and used gradient boosting for predicting the interview variable. Facial features and meta attributes extracted with SHORE\footnote{\url{https://www.iis.fraunhofer.de/en/ff/bsy/tech/bildanalyse/shore-gesichtsdetektion.html}} were used as visual descriptors. Pitch and intensity attributes were extracted from the audio signal. Finally, hand picked terms were used from the ASR transcriptions. The three type of features were concatenated and gradient boosting regression was applied for predicting traits and interview variable. 

\item \textbf{FDMB.} Used frame differences and appearance descriptors at multiple fixed image regions with a SVR method for predicting the interview variable and the five personality traits. After face detection and normalization, differences between consecutive frames was extracted. LPQ descriptors were extracted from each region in each frame and were concatenated. The video representation was obtained by adding  image-level descriptor. SVR was used to estimate traits and the interview variable. 
\end{itemize}

It was encouraging that the teams that completed the final phase of the first stage proposed methods that relied on diverse and complementary features and learning procedures. In fact, it is quite interesting that solutions based on deep learning were not that popular for this stage. This is in contrast with previous challenges in most aspects of computer vision (see e.g.~\cite{clapresources}), including the first impressions challenge~\cite{lopez2016eccv,icpr_contest}. In terms of the information/modalities used, all participants considered visual information, through features derived from faces and even context.  Audio was also considered by two out of the four teams. Whereas ASR transcripts were used only by a single team. Finally, information fusion was performed at a feature level.


\subsubsection{Stage 2 results: Explaining recommendations}\label{sec:stage2results}
The two teams completing the final phase of the qualitative stage were BU-NKU and TUD, and their approaches were detailed in previous subsections. Other teams also developed solutions to the explainability track, but did not succeed in submitting predictions for the test videos. BU-NKU and TUD were tied for the first place in the second stage.

Table~\ref{tab:resultss2} shows the results of participants in the explainability stage of the challenge. Recall that a committee of experts evaluated a sample of videos labeled with each methodology, using the measures described in Section~\ref{subsec:evalmeas}, and a [0,5] scale was adopted. It can be seen from this table that both methods obtained comparable performance. BU-NKU outperformed clearly the TUD team in terms of perceived clarity and interpretability, whereas the opposite happened in terms of creativity. 
\begin{table}[h!tb]
   \caption{Results of the second stage of the job screening coopetition.}
	\label{tab:resultss2}
   \begin{center}
   	\resizebox{\columnwidth}{!}{%
       \begin{tabular}{cccccccc} 
       \hline
\textbf{Rank}&\textbf{Team}&\textbf{Clarity}&\textbf{Explainability}&\textbf{Soundness}&\textbf{Interpretability}&\textbf{Creativity}&\textbf{Mean score}\\\hline
1&BU-NKU&4.31&3.58&3.4&3.83&2.67&3.56\\ 
1&TUD&3.33&3.23&3.43&2.4&3.4&3.16\\\hline
       \end{tabular}
       }
   \end{center}
\end{table}

The performances in Table~\ref{tab:resultss2} illustrate that there is room for improvement for developing proper explanations. In particular, evaluation measures for explainability deserve further attention.



\subsection{Discussion}
This section described the design of the job candidate screening challenge, as well as its top performing submissions. The challenge comprised two stages, one of which focused entirely on generating explanations for the recommendations made by models. Out of 72 participating teams, only two teams successfully completed both stages, illustrating the difficulty in generating explanations for complex machine learning pipelines. 

The two solutions that were described in detail comprise different methodologies and focus on different aspects. The BU-NKU system focused mostly on visual information for explaining recommendations. Audio was used as a complementary feature. The TDU system, on the other hand, gave more priority to textual information. Both methodologies are quite interesting and surely will be the basis for further research in this growing topic. 
\section{Analysis of the First Impressions data set}
\label{sec:analisys_data}
The collected personality traits dataset is rich in terms of the number of videos and annotations, and hence suitable for training models with high generalization power. The ground truth annotations used in training models are those given by individuals and may reflect their bias/preconception towards the person in the video, even though it may be unintentional and subconscious. Thus, the classifiers trained can inherently contain this subjective bias.

\edition{In the next subsection we analyze different aspects of the First Impression database, such as the video split procedure used to build the dataset (Sec.~\ref{sec:videosplit}), the intra/inter-video labeling variation (Sec.~\ref{sec:labelingvariation}), and subjective bias with respect to gender, ethnicity (Sec.~\ref{sec:genderethnicity}) and age (Sec.~\ref{sec:ageanalysis}).} 
\edition{We finally handle the problem as a binary classification task, which provides more room for improvement using multimodal approaches (Sec.~\ref{sec:experimental_quant_cls}).}

\subsection{\edition{Video Split}}\label{sec:videosplit}

\edition{As briefly mentioned at the beginning of Section~\ref{sec:overview}, the 10k clips of the First Impression dataset were obtained by partitioning 3k videos obtained from YouTube into small clips of 15 seconds each. Some of these small clips, generated from the same video, can be found in the train, validation and test set. However, it should be noted that even though different clips generated fom the same video can be found in different sets, they were captured at different time intervals. Thus, both the scene as well as the person appearing on each clip can have high variation in appearance due to different views/poses, camera motion and behavior.}

\begin{figure}[thpb]
	\centering
	\subfigure[\{0.67, 0.81, 0.74, 0.53, 0.55, 0.63\}]{\includegraphics[height=2.0cm]{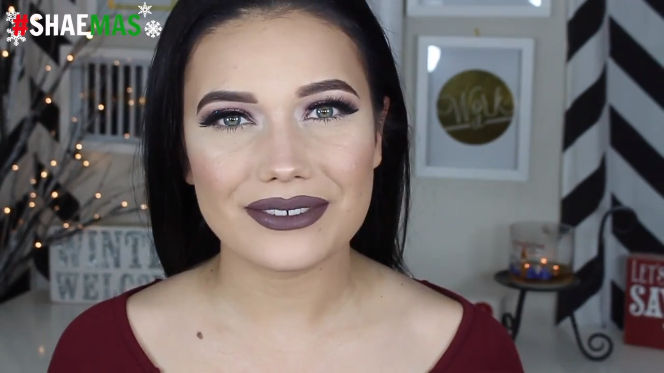}}\hspace{0.5cm}
	\subfigure[\{0.46, 0.62, 0.63, 0.44, 0.42, 0.47\}]{\includegraphics[height=2.0cm]{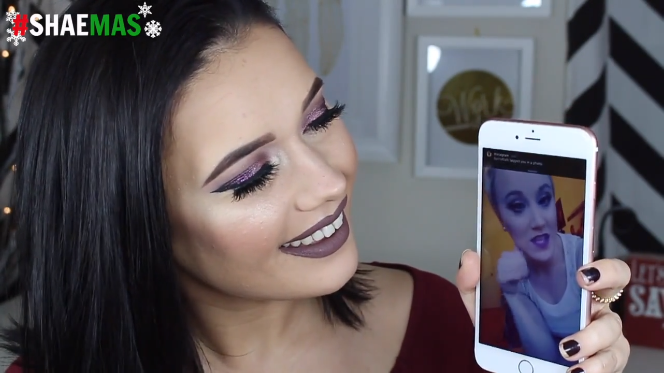}}
	\caption{\edition{Two frames extracted from different clips (generated from the same video) and respective labels \{Interview, O, C, E, A, N\}, which illustrate the intra-video variation in appearance/pose and labeling.}}
    \label{fig:intra-video}
\end{figure}

\edition{We analyzed the percentage of clips generated from the same video that are contained in different sets and results show that $83.7\%$ of the clips in the validation set have at least one clip in the train set which was generate from the same video. Similarly, $84\%$ of the clips in the test set have at least one clip in the train set which was generated from the same video. From the exact $3,060$ original videos, $721$ have not been split and $2,339$ have been split at least once. The average split per video for the whole dataset is $3.27$ ($\pm1.8$) and the maximum number of splits found is $6$. Figure~\ref{fig:intra-video} shows two frames obtained from the same video but at different splits and respective labels associated to each clip, to illustrate the intra-video labeling variation.} 

\subsection{\edition{Intra-Video and Inter-Video labeling variation}}\label{sec:labelingvariation}

\edition{In this section we analyze the labeling variation with respect to different clips generated from the same video (intra-video), as well as  different videos from the same YouTube user (inter-video). For the former, we compute the standard deviation per trait for each video which has at least \edition{one} split (i.e., number of clips $>$ 1). Then, we computed the average deviation per video, taking into account the deviation for all traits. Finally, given the average deviation per video, we compute the global average deviation with respect to the whole dataset. Results are shown in Table~\ref{inter:intra:videoanalysis}.}

\begin{table}[htbp]
\centering
\caption{\edition{Intra-Video and Inter-Video variation analysis. Second and third columns show the average standard deviation per video (intra-video) and per user (inter-video) for the 6 variable under analysis, i.e., \{Interview, O, C, E, A, N\}. ``Global avg deviation'' represents the mean standard deviation over all traits with respect to the whole dataset (following the intra/inter-video procedures described in this section).}}
\label{inter:intra:videoanalysis}
\begin{tabular}{ccc}
 \hline
               & \edition{\textbf{Intra-video}} & \edition{\textbf{Inter-video}} \\ \hline
\edition{Interview}      & \edition{0.064 ($\pm0.033$)} & \edition{0.054 ($\pm0.040$)} \\ 
\edition{O}              & \edition{0.070 ($\pm0.034$)} & \edition{0.059 ($\pm0.044$)} \\ 
\edition{C}              & \edition{0.063 ($\pm0.031$)} & \edition{0.055 ($\pm0.041$)} \\ 
\edition{E}              & \edition{0.068 ($\pm0.035$)} & \edition{0.056 ($\pm0.044$)} \\ 
\edition{A}              & \edition{0.071 ($\pm0.035$)} & \edition{0.048 ($\pm0.037$)} \\ 
\edition{N}              & \edition{0.071 ($\pm0.034$)} & \edition{0.052 ($\pm0.040$)} \\ 
\edition{Avg.} 			 & \edition{0.068 ($\pm0.024$)} & \edition{0.054 ($\pm0.032$)} \\ \hline
\end{tabular}
\end{table}

\edition{For the inter-video analysis, we first grouped all videos from each user (although some users might have videos from different people, we assume most videos of the same user are from the same individual, i.e., having the same individual appearing on them). First, consider the 10k videos in our dataset have been obtained from around 2762 YouTube users (i.e., 313 videos have no user associated with and are not included in this analysis). The average number of videos per user is $1.07$ ($\pm0.29$) and the maximum number of videos per user is $4$. The average number of clips (after split) per user is $3.51$ ($\pm2.12$) and the maximum number of clips per user is $16$. The procedure described next is performed for all users with number of videos $>1$ (i.e., $181$ users).} 

\edition{For each user, within these 181 users, we first computed the average label per trait for each video (assuming each video can be split into different clips). Then, we computed the standard deviation per trait taking into account the precomputed ``average labels'' with respect to all videos of a specific user. Then, we computed the average deviation over all traits with respect to each user, and finally, given the average deviation of all traits/users, we computed the global average deviation, as shown in the right side of Table~\ref{inter:intra:videoanalysis}.}

\begin{figure}[thpb]
	\centering
	\subfigure[\{0.67, 0.81, 0.74, 0.53, 0.55, 0.63\}]{\includegraphics[height=2.0cm]{figures/26CMYJn3u6Y_000.png}}\hspace{0.5cm}
	\subfigure[\{0.58, 0.79, 0.58, 0.66, 0.66, 0.62\}]{\includegraphics[height=2.0cm]{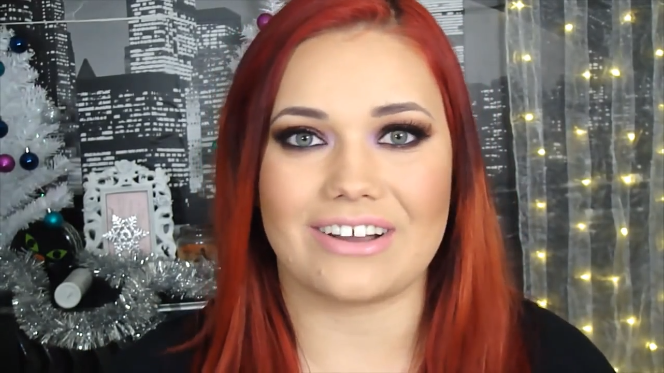}}\hspace{0.5cm}
	\subfigure[\{0.74, 0.83, 0.78, 0.62, 0.68, 0.64\}]{\includegraphics[height=2.0cm]{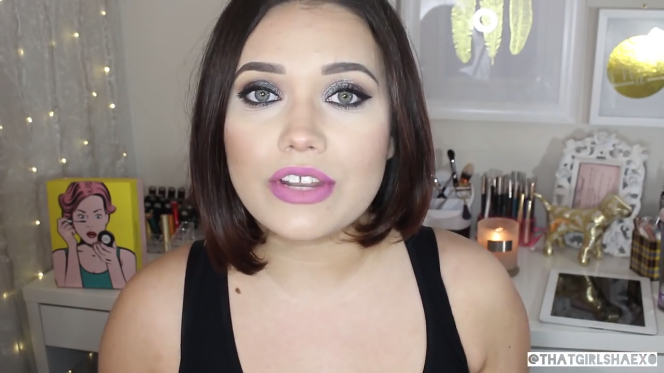}}    
	\caption{\edition{Three frames extracted from different videos of the same user (i.e., with the same person appearing on them) and respective labels \{Interview, O, C, E, A, N\}, which illustrate the inter-video variation in appearance and labeling.}}
    \label{fig:inter-video}
\end{figure}

\edition{It can be observed from Table~\ref{inter:intra:videoanalysis} that intra-video variation is higher than inter-video variation. This may happened because each set of pairs was annotated by a unique annotator. Thus, different clips of the same video may be annotated by different annotators. It can be also observed that there is not a global trend with respect to all analyzed traits and scenarios, i.e., the trait with lower/higher variation in the inter-video scenario may not have the lower/higher variation in the intra-video scheme. 
To  illustrate the inter-video variation, Figure~\ref{fig:inter-video} shows diferent frames, obtained from different videos of the same user, and respective labels.}

\subsection{Gender and Ethnicity analysis}\label{sec:genderethnicity}

In this section, existence of this latent bias towards gender\footnote{We follow the computer science literature here, and use ``gender estimation," but distinguishing male vs. female is more appropriately termed as ``sex estimation". Gender is a more complex and subjective construct.} and apparent ethnicity is analyzed. For this purpose, the videos used in the challenge are further manually annotated for gender and ethnicity, to complement the challenge meta data. Then a linear (Pearson) correlation analysis is carried out between these traits and apparent personality annotations. The results are summarized in Table~\ref{tab:gender_eth_bias}. Although the correlations range from weak to moderate, the statistical strength of the  relationships are very high. 

We first observe that there is an overall positive attitude/preconception towards females in both personality traits (except Agreeableness) and job interview invitation. The second observation is that the gender bias is stronger compared to ethnicity bias. Concerning the ethnicities, the results indicate an overall positive bias towards Caucasians, and a negative bias towards African-Americans. There is no discernible bias towards Asians in either way. 

\begin{table}[htbp]
  \centering
  \caption{Pearson correlations between annotations of gender-ethnicity versus personality traits and interview invitation. * and ** indicate significance of correlation with $p < 0.001$ and $p < 10^{-6}$, respectively.}
    \begin{tabular}{llllc}
    \hline
    \textbf{Correlation} &{\textbf{Gender}} & \multicolumn{3}{c}{\textbf{Ethnicity}} \\ \hline
    \textbf{Dimension} & {\textbf{Female}} & {\textbf{Asian}} & {\textbf{Caucasian}} & {\textbf{Afro-American}} \\ \hline
    $Agreeableness$ & -0.023 & -0.002 & \textbf{0.061**} & \textbf{-0.068**} \\ 
    $Conscientiousness$ & \textbf{0.081**} & 0.018 & \textbf{0.056**} & \textbf{-0.074**} \\ 
    $Extroversion$ & \textbf{0.207**} & \textbf{0.039*} & \textbf{0.039*} & \textbf{-0.068**} \\ 
    $\overline{Neuroticism}$ & \textbf{0.054*} & -0.002 & \textbf{0.047*} & \textbf{-0.053**} \\ 
    $Openness$ & \textbf{0.169**} & 0.010 & \textbf{0.083**} & \textbf{-0.100**} \\ 
    Interview & \textbf{0.069**} & 0.015 & \textbf{0.052*} & \textbf{-0.068**} \\\hline
    \end{tabular}%
  \label{tab:gender_eth_bias}%
\end{table}%

When correlations are analyzed closely, we see that women are perceived as more ``open" and ``extroverted'' compared to men, noting that the same but negated correlations apply for men. It is also seen that women have higher prior chances to be invited for a job interview. We observe a similar, but negative correlation with the apparent Afro-American ethnicity. To quantify these, we first measure the trait-wise means from the development set, comprised of 8000 videos. We then binarize the interview variable using the global mean score, and compute prior probability of job invitation conditioned on gender and ethnicity traits. The results summarized in Table~\ref{tab:ge_means_probs} clearly indicate a difference in the chances for males and females to be invited for a job interview. Furthermore, the conditional prior probabilities show that Asians have an even higher chance to be called for a job interview compared to Caucasian ethnicity, while Afro-Americans are disfavored. Since these biases are present in the annotations, supervised learning will result in systems with similar biases. Such algorithmic biases should be made explicit for preventing the misuse of automatic systems.

\begin{table}[htbp]
  \centering
  \caption{Gender and ethnicity based mean scores and conditional prior probabilities for job interview invitation.}
    \begin{tabular}{lccccc}
    \hline
          & \multicolumn{1}{l}{\textbf{Male}} & \multicolumn{1}{l}{\textbf{Female}} & \multicolumn{1}{l}{\textbf{Asian}} & \multicolumn{1}{l}{\textbf{Caucasian}} & \multicolumn{1}{l}{\textbf{Afro-American}} \\
    \hline
    mean scores & 0.539 & 0.589 & 0.515 & 0.507 & 0.475 \\
    
    p(invite $\vert$ trait) & 0.495 & 0.560 & 0.562 & 0.539 & 0.444 \\
    \hline
    \end{tabular}%
  \label{tab:ge_means_probs}%
\end{table}%

\subsection{Age analysis}\label{sec:ageanalysis}

\begin{figure}[htb] 
    \centering    
    \includegraphics[scale=0.4]{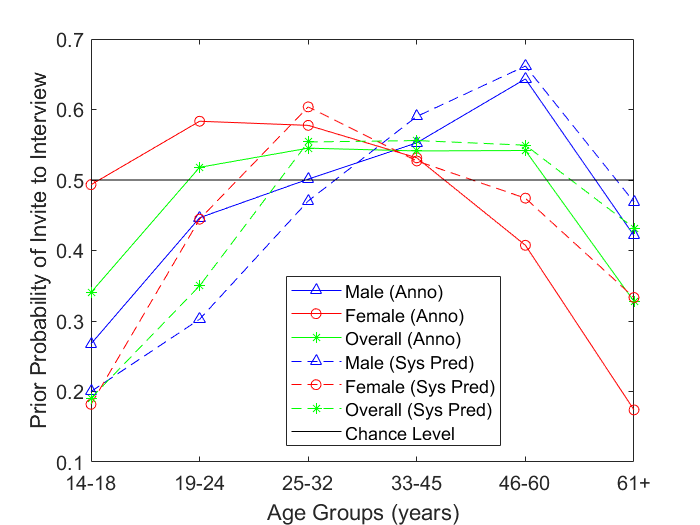} 
    \caption{The prior probability of job interview invitation over age groups, and over gender and age groups jointly.}\label{fig:age_gender_analysis}
\end{figure}

We annotated the subjects into eight disjoint age groups using the first image of each video. The people on the videos are classified into one of the following groups: 0-6, 7-13, 14-18, 19-24, 25-32, 33-45, 46-60 and 61+ years old. We excluded the 13 subjects under 14 years old from the analysis. We subsequently analyzed the prior probability of job interview invitation for each age group, with and without gender breakdown. \edition{Apart from the ground truth annotations, we also analyzed the held out test set predictions from BU-NKU (winner) system explained in Section~\ref{section:bunku}.} The results  are summarized in Figure~\ref{fig:age_gender_analysis}\edition{, where``(Anno)" refers to statistics from ground truth annotations and ``(Sys Pred)" are those from the test set predictions}.

\edition{Regarding the ground truth annotations,} on the overall, the prior probability is lower than 0.5 (chance level) for people under 19 or over 60. This is understandable, as very young or old people may not be seen (legally and/or physically) fit to work. For people whose age range from 19 to 60 (i.e. working-age groups), the invitation chance is slightly (but not significantly) higher than the chance level. Within the working-age groups, the female prior probability peaks at 19-24 age group and decreases with increasing age, while for the male gender the prior probability of job invitation steadily increases with age. The analysis shows that although non-consciously, people prefer to invite women when they are younger and men when they are older to a job interview. This is also verified with correlation analysis: the Pearson correlation between the ordinal age group labels and ground truth interview scores are 0.126 (p $< 10^{-13}$) and 0.074 (p $< 10^{-5}$) for male and female gender, respectively.  The results indicate that likability/fitness may be an underlying factor in female job invitation preference. For males, the preference may be attributed to the perceived experience and authority of the subject.

\edition{While the analysis of the ground truth reflects a latent bias towards age and gender combination, we see that the classifiers trained on these annotations implicitly model the bias patterns. For male candidates, the classifier exhibits the same age-gender bias pattern for job interview invitation compared to the ground truth and the statistics for males older than 25 years are very similar in both. A similar pattern exists for the female candidates however peaking at the age of 25-32, instead of 19-24. The largest difference between the ground truth and the classifier statistics is observed in non-working-age groups (14-18 and 61+) and particularly among females. This may be attributed to the fact that these groups form a small proportion (3.5\%) of the data.} 

The analysis provided in \edition{Sections~\ref{sec:genderethnicity} and~\ref{sec:ageanalysis}}  evidences potential biases for systems trained on the first impressions data set we released. Therefore, even when the annotation procedure aimed to be objective, biases are difficult to avoid. Explainability could be an effective way to overcome data biases, or at least to point out these potential biases so that decision takers can take them into account. Also, please note that explainable mechanisms could use data-bias information to provide explanations on their recommendations. 
\subsection{Handling the Problem as a Classification Task}
\label{sec:experimental_quant_cls}
\begin{figure}
	\centering
	\includegraphics[width=.66\textwidth]{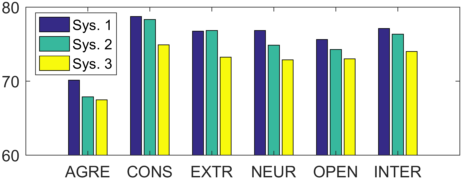}
	\caption{\edition{Test set classification performance of top three fusion systems over personality traits and the interview variable. Sys. 1: Audio-Video system, Sys. 2: Video only system, Sys. 3: Audio plus a single image based system. NEUR refers to non-Neuroticism as it is used throughout the paper.}}
	\label{fig:top3_cls_sys}
\end{figure}

\edition{
The results and analysis presented in Section~\ref{subsec:stage1results}  suggested the values of most of the considered variables lied close to the mean. With the goal of bringing more light on the difficulty of the problem, in this section we report preliminary results when approaching the problem as one of classification instead of regression.  
For this experiment we considered the   
BU-NKU system described in Section~\ref{sec:solutions_from_participants}. Additionally, we analyze how well a parsimonious system can do by looking at a single frame of the video, instead of face analysis in all frames.

To adapt the problem for classification, the continuous target variables in the [0,1] range are binarized using the training set mean statistic for each target dimension, separately.
For the single-image tests, we extracted deep facial features from our fine-tuned VGG-FER DCNN, and accompanied them with easy-to-extract image descriptors, such as Local Binary Patterns (LBP)~\cite{ojala2002multiresolution}, Histogram of Oriented Gradients (HOG)~\cite{dalal2005histograms} and Scale Invariant Feature Transform~\cite{lowe2004distinctive}.
Hyper-parameter optimization and testing follow similar schemes as in Section~\ref{section:bu-nku-quan}. The test set classification performances of the top systems for single- and multi-modal approaches are shown in Table~\ref{tab:classification}. 

First of all, it can be noticed that the performance of the classification model lies between 69\% and 77\% of test set accuracy. Since the values of traits follow a nearly symmetric distribution (see Figures~\ref{hist:datasettrain} and~\ref{hist:datasettest}), random guessing would approximate a 50\% of accuracy. This illustrates the difficulty of the classification task and suggests that the apparently low gap to reach perfect performance in the regression problem is far from being reached. 

As expected, we see that the audio-visual approach also performs best in the classification task (77.10\% accuracy on the interview variable). This is followed by the video-only approach using facial features (76.35\%), and the fusion of audio with face and scene features from the first image (74\%). Although this is relatively 4.6\% lower compared to the best audio-visual approach, it is highly motivating, as it uses only a single image frame to predict the personality impressions and interview invitation decision, which the annotators gave by watching the whole video. It shows that without resorting to costly image processing and DCNN feature extraction for all images in a video, it is possible to achieve high accuracy, comparable to the state-of-the-art. 
 
The dimension that is the hardest to classify is agreeableness, whereas accuracy for conscientiousness was consistently the highest (see Figure~\ref{fig:top3_cls_sys}). Among the conventional image descriptors, HOG was the most successful, with an average validation set recognition accuracy (over traits) of 70\%, using only a single facial image. On the other hand, the fusion of scene and face features from the first video frame outperforms acoustic features on both the development and test sets by 3\%. }

\begin{table}[htbp]
  \centering
  \caption{\edition{Test set classification accuracies for the top single and multimodal systems. The scene feature is extracted from the first video frame.} }
   	\begin{tabular}{llcc}
   	\hline
	\edition{\textbf{Sys.}}	&\edition{\textbf{Modality}} &\edition{\textbf{Interview}} & \edition{\textbf{Trait Avg.}} \\
		\hline
		\edition{1} &\edition{Audio + Video }&  \edition{77.10} & \edition{75.63} \\
		\edition{2} &\edition{Video (Face Seq.)} & \edition{76.35} & \edition{74.45} \\
		\edition{3} &\edition{Audio + Scene + First Face}  & \edition{74.00} & \edition{72.31} \\
		\edition{4} &\edition{Audio + Scene} & \edition{71.95} & \edition{70.47} \\
		\edition{5} &\edition{First Face + Scene}  & \edition{71.15} & \edition{69.97} \\
		\edition{6} &\edition{Audio Only} & \edition{69.25} & \edition{67.93} \\
		\hline
	\end{tabular}%
  \label{tab:classification}%
\end{table}%


\section{Lessons learned and open issues}
\label{sec:lessons_learned}
Explainable decision making is particularly important for algorithmic discrimination, where people are affected by the decisions given by an algorithm~\cite{freuler}. Such algorithms may be used for prioritization (e.g. multimedia search engines), classification (e.g. credit scoring), association (e.g. predictive policing), or filtering (e.g. recommender systems). In this paper, we have described the first comprehensive challenge on apparent personality estimation, proposed two end-to-end solutions, and investigated issues of algorithmic accountability. 

The first thing we would like to stress is that explainability requires user studies for its evaluation. Testbeds and protocols, such as the one we contribute in this paper, will be useful for advancing research in explainability. Additionally, it is essential that algorithmic accountability is broken down into multiple dimensions, along which systems are evaluated. In~\cite{diakopoulos}, these dimensions are proposed as responsibility, explainability, accuracy, auditability, and fairness, respectively. We have proposed five dimensions for explainability in this work. 

Several applications and domains within computer vision will benefit from having explainable and interpretable models. Such mechanisms are essential in scenarios in which the outcome of the model can have serious implications. We foresee that the following application domains will significantly benefit from research in this direction: health applications (e.g., model-assisted diagnosis, remote patient monitoring, etc.); \emph{non-visually-obvious} human behavior analysis (e.g., personality analysis, job screening); recognition tasks involving people (e.g., gender, ethnicity, age  recognition); cultural-dependent tasks (e.g., adult content classification, cultural event recognition); security applications (e.g., biometrics of potential offenders, detection/verification/scanning systems, smart surveillance, etc.). The availability of new explicable/interpretable models will increase the scope of research for computer vision problems, and allow the creation of human-computer mixed decision systems in more sensitive application areas.

There is a marked distinction between visual question answering (VQA) and explainable decision making. VQA produces a narrative of the  multimedia input, whereas explainability requires making a narrative of the decision process itself. This can be seen as a meta-cognitive property of the system. At the moment, the focus is on natural language based explanations, as well as strong visualizations suitable for human interpretation. Once such systems are sufficiently advanced, we could expect machine-interpretable explanations (such as through micro ontologies) to be produced as by-products, and compartmentalized systems taking advantage of such explanations to improve their decision making. 

Black box models, such as deep neural network approaches, require external mechanisms (such as systematic examination of internal responses for ranges of input conditions) for the interpretation of their workings, which increase the annotation and training burden of these systems. On the other hand, transparent (white) models trade off accuracy. Balanced systems, such as the solutions we proposed in this work, combining early black box modeling with transparent decision-level modeling, could be the ideal solution.

\edition{Considering the current scenario of first-impression analysis in the context of job candidate screening, the quest for algorithmic accountability will inevitably also bring ethical questions. Will automatically trained pipelines inadvertently pick up on data properties that actually are irrelevant to the problem at hand~\cite{Sturm2014Ahorse}? May certain individuals get disadvantaged because of this, or due to inherent data biases? Generally, can judgments of external assessors be trusted? These are challenging questions, that will need comprehensive, interdisciplinary effort in order to be answered. In~\cite{Liemetal2018}, an extensive discussion of this topic is given, considering the algorithmic job candidate screening problem from a machine learning and organizational psychology perspective.}

\edition{Through the explicit focus of the ChaLearn challenge on \emph{apparent} personality analysis, the question on whether prediction labels reflect `true' personality is not the central question. Instead, the focus lies on inferring potential patterns in first-impression judgments made by outsiders. While these may be biased, the goal at this stage is not to mitigate this bias, but rather to gain deeper insight into what data characteristics may be underlying observed biases.}

\edition{Contrasting typical approaches in organizational psychology with the approach taken in the ChaLearn challenge, traditional psychometrically validated personality measurement instruments would involve many more item questions than the ones that were currently offered to MTurk workers. In addition, vlog data in the challenge may not originally have been intended as a video resume for a professional job candidacy. At the same time, the current data acquisition setup allowed for considerable scaling in terms of the amount of videos that could be annotated and analyzed; while several dozens of video resumes may already be a considerably sized corpus in the psychology domain, in the current challenge, several thousands of clips were studied.}

\edition{These very different approaches therefore can offer different viewpoints and insights on the problem. A next step for future work will be to more explicitly compare them. Generally, in case inherent data and judgment bias would lead to unfair or unethical system predictions, the responsibility for avoiding this is shared. Machine learning researchers and engineers should be aware of this, and can conceive algorithmic solutions that explicitly mitigate unethical outcomes. At the same time, the identification of the most critical ethical challenges also needs explicit input and insight from data providers and domain specialists.}

\begin{acknowledgements}

The challenge organizers gratefully acknowledge a grant from Azure for Research, which allowed running the challenge on the Codalab platform and the technical support of Universit\'e Paris-Saclay. ChaLearn provided prizes and travel awards to the winners.
This work was partially supported by CONACyT under grant 241306, Spanish Ministry projects TIN2016-74946-P and TIN2015-66951-C2-2-R (MINECO/FEDER, UE), and CERCA Programme / Generalitat de Catalunya. H.J. Escalante was supported by \emph{Red Tem\'aticas CONACyTs en Tecnolog\'ias del Lenguaje (RedTTL) e Inteligencia Computacional Aplicada (RedICA)}. A.A. Salah was supported by the BAGEP Award of the Science Academy. We gratefully acknowledge the support of NVIDIA Corporation with the donation of the Titan Xp GPU used for this research. 

\end{acknowledgements}



\end{document}